\documentclass{article}

\usepackage[utf8]{inputenc} 
\usepackage[T1]{fontenc}    
\usepackage{hyperref}       
\usepackage{url}            
\usepackage{booktabs}       
\usepackage{amsfonts}       
\usepackage{nicefrac}       
\usepackage{microtype}      
\usepackage{xcolor}         

\usepackage{amsmath}
\usepackage{graphicx}
\usepackage{natbib}
\usepackage{subcaption}

\usepackage{mathtools}
\usepackage{stmaryrd}

\title{Improving the Validity of Decision Trees \\ as Explanations}

\author{%
  Jiri Nemecek, Tomas Pevny, and Jakub Marecek \\
  \footnotesize
  Department of Computer Science\\
  \footnotesize Czech Technical University in Prague\\
  \footnotesize \texttt{contact@nemecekjiri.cz}
}

\begin{document}

\maketitle

\begin{abstract}
    In classification and forecasting with tabular data, one often utilizes tree-based models. 
  Those can be competitive with deep neural networks on tabular data and, under some conditions, explainable. 
  The explainability depends on the depth of the tree and the accuracy in each leaf of the tree. 
  We point out that decision trees containing leaves with unbalanced accuracy can provide misleading explanations. Low-accuracy leaves give less valid explanations, which could be interpreted as unfairness among subgroups utilizing these explanations. 
  Here, we train a shallow tree with the objective of minimizing the maximum misclassification error across all leaf nodes.
  The shallow tree provides a global explanation, while the overall statistical performance of the shallow tree can become comparable to state-of-the-art methods (e.g., well-tuned XGBoost) by extending the leaves with further models. 
\end{abstract}

\section{Introduction}



In classification and forecasting with tabular data, one often utilizes axis-aligned decision trees \cite{Payne1977AnAF,breiman1984classification}. 
  A prime example of a high-risk application of AI, where decision trees are widely used, is credit risk scoring \cite{mays1995handbook,lessmann2015benchmarking,thomas2017credit} in the financial services industry \cite{athey2019machine}.
  There, the relevant regulation, such as the Equal Credit Opportunity Act in the US \cite{ECOA} and related regulation \cite{CreditDir,GDPR} in the European Union, bars the use of models that are not explainable \cite{rudin2019stop}, 
  which is often construed \cite{BritishDiscussionPaper,FrenchDiscussionPaper,gunnarsson2021deep,Circular} as requiring the use of decision trees.
  When studying the decision tree that a bank uses, one often focuses on ways that would make it possible to obtain a loan,
  and one would wish that the corresponding leaf of the decision tree had as high accuracy as possible.  

  In many other domains, the use of tree-based models has a long tradition.
  Consider, for example, judicial applications of AI such as the infamous Correctional Offender Management Profiling for Alternative Sanctions (COMPAS) \cite{brennan2009evaluating,Brennan2018,courtland2018bias,Zhou2023}, which is marketed as the ``nationally recognized decision tree model'', 
  or medical applications of AI \citep[e.g.,][]{rakha2014nottingham,london2019artificial,tjoa2020survey}. 
  It is hard to overstate the importance of high accuracy of any rule that a medical doctor or a judge may learn from a decision tree. 
  Decision trees are also used in model extraction \cite{bastani2017interpreting} to provide globally valid explanations of black-box classifiers.

  Shallow trees can indeed serve as global explanations for a classifier---or explainable classifiers \emph{per se}---when each leaf is construed as a logical rule. 
  Because various individuals or subgroups may deem various outcomes of importance, 
a \emph{fair explanation} would have as high accuracy in each leaf of the decision tree as possible.  
  Additionally, the depth needs to be low\footnote{According to~\citet{feldman2000minimization}, humans can understand logical rules with boolean complexity of up to 5--9, depending on their ability, where the boolean complexity is the length of the shortest Boolean formula logically equivalent to the concept, usually expressed in terms of the number of literals.} in order for the rule explaining the decision in each leaf to remain comprehensible.  

\begin{figure}
  \centering
  \includegraphics[width=0.75\linewidth]{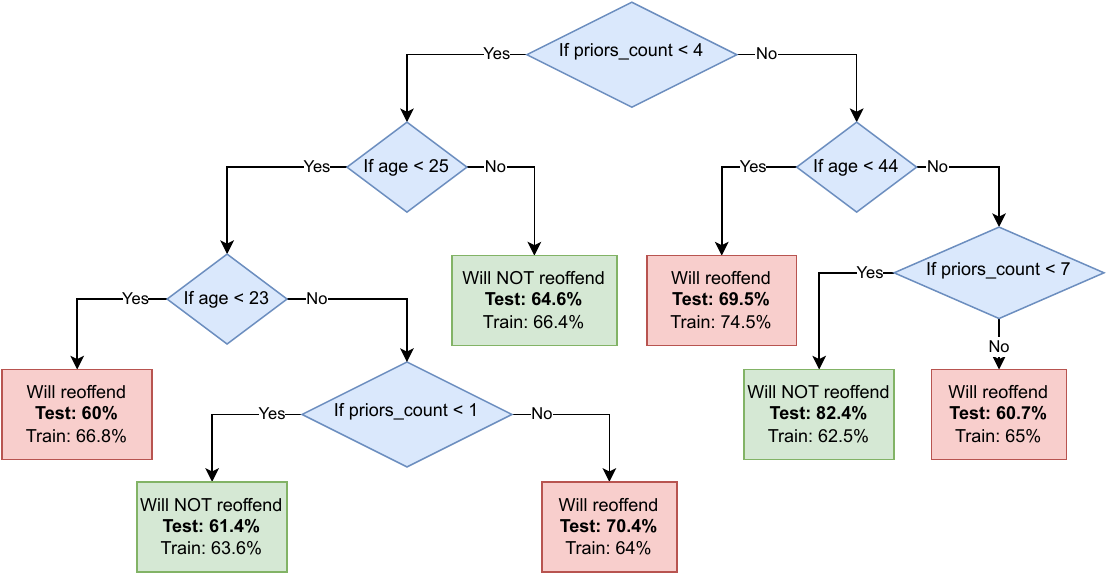}
  \caption{An example of the decision tree produced by the proposed model for the COMPAS dataset, cf. Figure \ref{fig:motiv}. The bold percentage shows the leaf accuracy in each leaf on out-of-sample data before applying the extending model. Below that, in regular font, we provide accuracy on training data. } 
    \label{fig:example}
\end{figure}

\begin{figure*}[t]
  \centering
  \includegraphics[width=\textwidth]{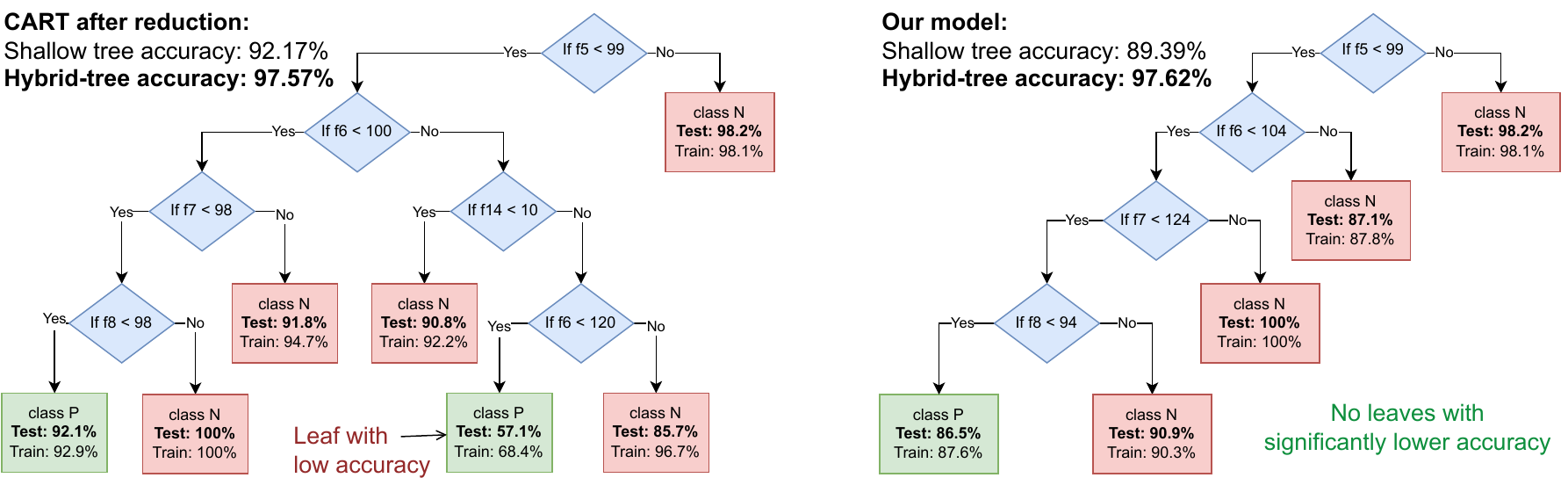} 
  \caption{A comparison of decision trees produced by CART and our method for \textit{pol} dataset. In each leaf, bold/regular percentage shows the leaf accuracy before extending it further on the test/training data set, respectively. Below the name of the model, we present the (hybrid-tree) accuracy of the hybrid/shallow tree in bold/regular font. 
  The CART tree contains a leaf with a notably lower accuracy compared to the overall accuracy of the model. The explanation provided by this leaf is less valid. This makes the global explanation provided by the tree less fair. While model accuracies do not take this into account, the proposed measure of leaf accuracy does. The left and right trees have leaf accuracy on unseen data equal 57.1\% and 86.5\%, respectively. 
  }
    \label{fig:pol_example}
\end{figure*}

  Similarly, one could argue that a decision tree can provide misleading explanations. 
  To evaluate how valid
or misleading the decision tree is, we suggest considering the minimal accuracy in any leaf of a tree (tree's \emph{leaf accuracy}).
  Indeed, a member of the public, when presented with the decision tree, may assume that each leaf of a decision tree can be construed as a logical rule.
  Consider, for example, the decision tree of Figure \ref{fig:example},
  based on the two-year variant of the well-known COMPAS \cite{brennan2009evaluating} dataset, which
  considers the binary classification problem of whether the individual would re-offend 
  within the next two years.
  The left-most leaf may be interpreted as suggesting that for up to 3 prior counts and under 23 years of age, the defendant will re-offend within the next two years after release. 
  However, the validity of this rule is somewhat questionable: the training accuracy in that leaf is 66.8\%, while the test accuracy is 60\% in that leaf.
  This suggests that 40\% of defendants who meet these criteria will actually not re-offend within two years.
  For a more extreme example, see 
  Figure~\ref{fig:pol_example}, which shows two trees of similar overall accuracy for the pol(e) dataset.
  When optimizing for overall accuracy, the minimum test accuracy in one leaf can be as low as 
  57.1\% (cf. the left tree in Figure~\ref{fig:pol_example}). 
  However, when maximizing the minimum training accuracy in one leaf,
  the minimum test accuracy in one leaf increases to 86.5\% (cf. the right tree in Figure~\ref{fig:pol_example}).
  One could argue that this improves the validity and fairness of the explanation provided by the tree.

  Although a recent comparison of the statistical performance of gradient-boosted trees and deep neural networks 
  by \citet{grinsztajnWhyTreebasedModels2022} has shown that the state-of-the-art tree-based models can outperform state-of-the-art neural networks across a comprehensive benchmark of tabular data sets, for our decision trees, the explainability requirement limits the overall accuracy.
To create a model with comparable total accuracy, we follow the existing works that combine interpretable models with black boxes, aiming for a (tunable) balance between accuracy and explainability \cite[e.g.,][]{wangGainingFreeLowcost2019,frostPartiallyInterpretableModels2024}. We can extend each of the leaves of the tree with a separate model, in our case, XGBoost. We have a valid explanation when the XGBoost model agrees with the tree, which is the majority of the cases. 
Our approach suggests an analogy to PCA, where one often interprets just a few components, leaving the rest uninterpreted, using them only to improve the statistical performance. Similarly, we train a well-balanced shallow tree that interprets the majority of the data and the remainder is estimated by XGBoost models. 
  
  In our approach, we use mixed-integer optimization (MIO) to train a shallow tree with the objective of minimizing the maximum misclassification error within each leaf.
  Seen another way, we maximize the minimal accuracy in any leaf of a tree.  
  In the second (optional) step,   
  we can train further models that extend each leaf of the shallow tree. 
  The shallow tree with the additional constraints on the accuracy in the leaves provides a fair interpretation, 
  while the overall accuracy of the hybrid trees \cite{zhou2002hybrid} combining shallow trees and the extending models (which we call the \emph{hybrid-tree accuracy}) improves upon (hybrid) decision trees trained using classical methods (e.g., CART) and is comparable to state-of-the-art tree-based methods,  
  such as the well-tuned XGBoost of \citet{grinsztajnWhyTreebasedModels2022}.

\begin{figure}[t]
  \centering
  \includegraphics[width=0.7\linewidth]{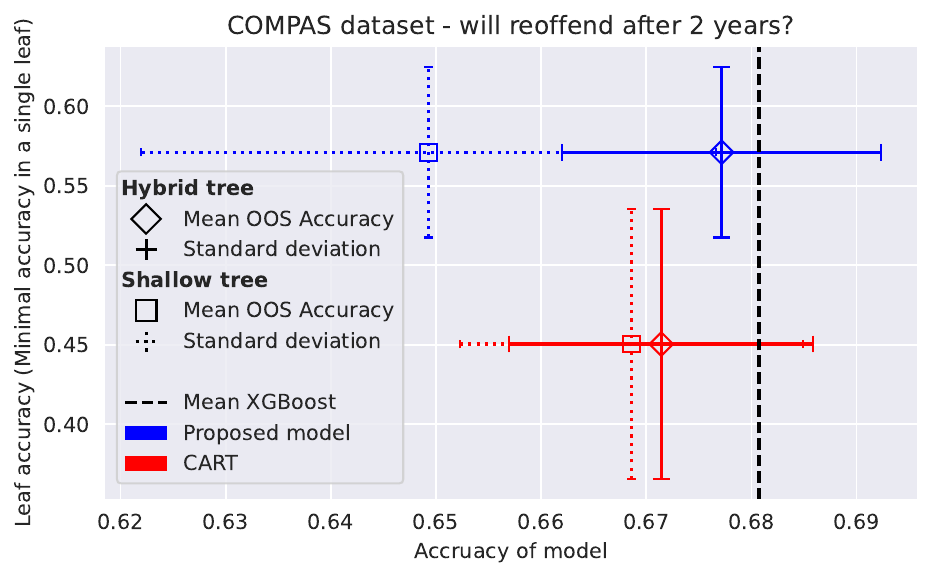}
  \caption{Performance on the COMPAS dataset: 
  Mean statistical performance 
  over 10 different train-test splits, evaluated in terms of model accuracy (horizontal axis) and leaf accuracy (vertical axis) for five variants of (hybrid) decision trees.
  The horizontal and vertical error bars are standard deviations across the 10 random runs.
  Notice that the proposed model has better interpretability compared to any standard decision tree and, once extended, accuracy comparable to a gradient-boosted tree.
 }
    \label{fig:motiv}
\end{figure}

  Let us illustrate the statistical performance. 
  Figure \ref{fig:motiv} shows that the accuracy of well-tuned XGBoost of \citeauthor{grinsztajnWhyTreebasedModels2022} on the two-year COMPAS \cite{brennan2009evaluating} test case exceeds 0.68.
  The accuracy of our shallow tree trained with the leaf-accuracy objective is below 0.65,
  which should not be surprising, considering the overall model 
  accuracy is \emph{not} the main objective.
    Nevertheless, by extending models in leaf nodes of the shallow tree, 
  we can improve the accuracy very close to 0.68,
  which improves upon both the accuracy of CART of the same depth alone (below 0.67) and 
  CART of the same depth undergoing the same leaf-extension procedure (slightly above 0.67).
  This performance is rather typical across the benchmark of \citeauthor{grinsztajnWhyTreebasedModels2022}. The proposed method outperforms CART with statistical significance, as detailed in Section \ref{sec:results}.
  
  \textbf{Our contributions.} We present:
  \begin{itemize}
  \item the challenge of \emph{fairness} (or, equivalently, \emph{validity}) of an explanation.
  \item leaf accuracy as a criterion for evaluating the validity and fairness of a classification tree as a global explanation. Leaf accuracy of decision tree $T$ is defined as follows:
  \begin{equation}
    A_L(T) \coloneqq \min_{l \in \mathcal{L}(T)} \frac{1}{|X_l|} \sum_{(\boldsymbol{x}, y) \in X_l} \llbracket y = C_l \rrbracket  \label{eq:leaf_acc}
  \end{equation}
  where $ \mathcal{L}(T)$ is the set of leaf nodes of tree $T$, $X_l \subseteq X$ is the set of samples assigned to the leaf $l$, and $C_l$ is the decision class of the leaf $l$ and $\llbracket a \rrbracket$ equals one if $a$ is true and zero otherwise. 
  \item a method for training decision trees that are optimal with respect to leaf accuracy, 
  which is scalable across a well-known benchmark \cite{grinsztajnWhyTreebasedModels2022},
  despite its use of mixed-integer optimization.
  \item benchmarking on tabular datasets \cite{grinsztajnWhyTreebasedModels2022} suggesting that the leaf accuracy can be significantly improved by up to 18.41 percentage points,
  while suffering only a very modest drop (at most 
  2.76 percentage points across the benchmark) in overall model accuracy, compared to well-tuned XGBoost \cite{grinsztajnWhyTreebasedModels2022}, once the shallow tree is extended.  
  \end{itemize}

\section{Related Work}

Decision trees \cite{breiman1984classification} 
are among the leading supervised machine learning methods, where interpretability and out-of-sample classification performance is important.
Random forests \cite{breiman2001random} and gradient-boosting tree-ensemble approaches \cite{mason1999boosting,Friedman2001} improve upon
their statistical performance substantially while limiting the interpretability. However, partially interpretable models \cite[e.g.][]{wangGainingFreeLowcost2019,frostPartiallyInterpretableModels2024} that combine an interpretable model with a black-box model have been shown to sometimes even improve the performance over pure black-box models \cite{lePraisePartiallyInterpretable2020}.

We are given $n$ samples $(x_{i1}, \ldots, x_{ip} ,y_i)$ with $p$ features each, for $i = 1 \ldots n$,
and their classification $y_i \in [K]$ into $K$ classes.
Let us denote sample $i$ by $\boldsymbol{x}_i = (x_{i1}, \ldots, x_{ip})$.
A decision tree sequentially splits a set of samples into two partitions:
In each non-leaf node $t$, it splits the samples based on their values $x_{:j_t}$ of a particular feature $j_t \in [p]$ and a threshold $b_t$. 
(See Figure~\ref{fig:example} for an illustration.)
More recently, decision trees have played an important role in explainable artificial intelligence \cite{arrieta2020explainable,burkart2021survey}
and interpretable machine learning \cite{rudin2022interpretable}.

Construction of an optimal axis-aligned binary decision tree is NP-Hard \cite{laurent1976constructing},
and hence all known polynomial-time algorithms, such as CART \cite{breiman1984classification}, produce suboptimal results, at least for some cases. 
Still, CART \cite{breiman1984classification}, which utilizes the Gini diversity index and cross-validation in pruning trees,
ranks among the leading algorithms \cite{wu2008top} in machine learning. 
A decade later, Breiman suggested that boosting can be interpreted as an optimization algorithm \cite{breiman1998arcing},
leading to the development of gradient-boosted trees \citep[e.g.,][]{mason1999boosting,Friedman2001}.
Their well-tuned variants \citep[e.g.,][]{Tianqi2016,NIPS2017_6449f44a,prokhorenkova2018catboost} are state-of-the-art polynomial-time algorithms for training decision trees. 
We refer to \cite{NEURIPS2021_9d86d83f,grinsztajnWhyTreebasedModels2022} for comparisons against deep neural networks. 

\citet{bertsimasOptimalClassificationTrees2017} and, independently, others \cite{Bessiere2009,Narodytska2018,gunluk2021optimal}, pioneered the use of exponential-time algorithms in the construction of decision trees.
The MIO formulation of \citeauthor{bertsimasOptimalClassificationTrees2017} suffers from 
some issues of scalability \cite{Verwer_Zhang_2019},
but can be easily extended by the addition of further constraints, such as sparsity \cite{hu2019optimal,xin2022exploring,ZhangEtAlOSRT2023}, fairness \cite{Verwer_Zhang_2019,van2022fair}, upper bounds on the number of leaves \cite{lin2020generalized}, incremental progress bounds \cite{lin2020generalized}, 
bounds on similarity of the support \cite{lin2020generalized}, 
a wide variety of privacy-related constraints, 
and in our case, 
accuracy in the leaves. 
Likewise, there are numerous extensions in terms of the objective \cite{lin2020generalized}, including 
F-score, AUC, and partial area under the ROC convex hull and, in our case, 
the leaf accuracy.
Subsequently, the \emph{optimal decision trees} have grown into a substantial subfield within machine learning research. 

There have been several important proposals of alternative convex-optimization relaxations for optimal decision trees: 
\citet{dash2018boolean} have demonstrated the use of an extended formulation in a column-generation (branch-and-price) approach;
\citet{zhu2020scalable} have introduced another alternative formulation and a number of valid inequalities (cuts);
\citet{aghaei2020learning} have introduced yet another alternative formulation based on the maximum flow problem. 
Independently, \citet{carreira2018alternating} suggested using non-linear optimization techniques, such as alternating minimization 
leading to much further research \cite{zantedeschi2021learning}. 
We refer to \citet{carrizosa2021mathematical,Nanfack2022} for overviews of mathematical optimization in the construction of decision trees. 

Much recent research \citep[e.g.,][]{pmlr-v119-vidal20a,demirovic2022murtree,van2022fair,hua2022scalable,Mazumder2022} has also focussed on improving the scalability 
of exponential-time algorithms for optimal decision trees by using branch-and-bound methods without relaxations in the form of 
convex optimization and, more broadly, dynamic programming. These approaches are sometimes seen 
as less transparent, as the mixed-integer formulation needs to be translated to the appropriate pruning rules or cost-to-go functions, which are
less succinct, and the correctness of the translation can be non-trivial to verify. 
Nevertheless, \citet{hua2022scalable} have demonstrated the scalability of their method to a dataset with over 245,000 samples 
(utilizing less than 2000 core-hours), for example. On a benchmark of 21 datasets from the UCI Repository with over 7,000 samples, 
the algorithm can improve training accuracy by 3.6\% and testing accuracy by 2.8\% compared to the current state-of-the-art.
This seems to validate the practical relevance of optimal decision trees.

\section{Mixed-Integer Formulation}
\label{sec:mip}
Mixed-Integer (Linear) Optimization (MIO) is a method of mathematical optimization similar to Linear Programming, with some of its variables limited to integer values. The goal is to maximize an objective function while satisfying a number of (linear) non-strict inequality constraints \cite{wolseyIntegerProgramming2021}.
Because of the global optimization capabilities, MIO enables our approach to not suffer from issues created by greedy top-down approaches like CART (e.g., Figure \ref{fig:pol_example}). 

We build upon the MIO formulation of \emph{optimal decision trees} \cite{bertsimasOptimalClassificationTrees2017}, changing the objective and adding novel constraints. Figure \ref{eq:completeFormulation} presents the entire MIO formulation.

\begin{figure}
    {\footnotesize
    \begin{align}
        \mathcolor{purple}{\max \; & Q \label{eq:objective} \\
        \textrm{s. t. } & Q \le \sum_{i=1}^n{S_{it}} + (1 - l_t) && \forall t \in \mathcal{T}_L \label{eq:Qltsum}\\
        & s_{it} \le z_{it} && \forall i \in [n], \; \forall t \in \mathcal{T}_L \label{eq:amount0}\\
        & r_{t} \le s_{it} + (1 - z_{it}) && \forall i \in [n], \; \forall t \in \mathcal{T}_L \label{eq:acref1}\\
        & r_{t} \ge s_{it} + (z_{it} - 1) && \forall i \in [n], \; \forall t \in \mathcal{T}_L \label{eq:acref2}\\
        & l_t = \sum_{i=1}^n{s_{it}} && \forall t \in \mathcal{T}_L \label{eq:sum1}\\
        & S_{it} \le s_{it} && \forall i \in [n], \; \forall t \in \mathcal{T}_L \label{eq:accequal1}\\
        & S_{it} \le \sum_{k=1}^K{Y_{ik} c_{kt}} && \forall i \in [n], \; \forall t \in \mathcal{T}_L \label{eq:misclas0}\\
        & S_{it} \ge s_{it} + \sum_{k=1}^K{Y_{ik} c_{kt}} - 1 && \forall i \in [n], \; \forall t \in \mathcal{T}_L \label{eq:accequal2} \\ 
        }& l_{t} = \sum_{k=1}^{K}{c_{kt}} && \forall t \in \mathcal{T}_L \label{eq:orig1} \\
        & \boldsymbol{a}_m^\mathsf{T} \boldsymbol{x}_i \ge b_m - (1 - z_{it}) && \forall i \in [n], \; \forall t \in \mathcal{T}_L, \nonumber\\
        & && \quad \qquad \forall m \in A_R(t) \label{eq:decr}\\
        & \boldsymbol{a}_m^\mathsf{T} (\boldsymbol{x}_i + \boldsymbol{\epsilon}) \le & & \forall i \in [n], \; \forall t \in \mathcal{T}_L, \nonumber \\
        & \; \quad b_m + (1 + \epsilon_{\max})(1 - z_{it}) && \quad \qquad \forall m \in A_L(t) \label{eq:decl}\\
        & \sum_{t \in \mathcal{T}_L}{z_{it}} = 1 && \forall i \in [n] \\
        & z_{it} \le l_t && \forall i \in [n], \; \forall t \in \mathcal{T}_L \\
        & \sum_{i=1}^n{z_{it}} \ge N_{\min} l_t && \forall t \in \mathcal{T}_L \label{eq:minN}\\
        & \sum_{j=1}^p{a_{jt}} = 1 && \forall t \in \mathcal{T}_B \\
        & 0 \le b_t \le 1 && \forall t \in \mathcal{T}_B \\
        & z_{it}, l_{t} \in \{0, 1\} && \forall i \in [n], \; \forall t \in \mathcal{T}_L \\
        & a_{jt} \in \{0, 1\} && \forall j \in [p], \; \forall t \in \mathcal{T}_B \\
        & c_{kt} \in \{0, 1\} && \forall k \in [K], \; \forall t \in \mathcal{T}_L \label{eq:origX} \\
        \mathcolor{purple}{& 0 \le Q, r_t, S_{it}, s_{it} \le 1 && \forall i \in [n], \; \forall t \in \mathcal{T}_L \label{eq:constrs}\\}
        \nonumber
    \end{align}
    }
\caption{The complete MIO formulation of training the shallow tree, maximizing the minimum accuracy across all leaf nodes, and constraining the number of samples per leaf node.
The constraints (\ref{eq:orig1} -- \ref{eq:origX}) are taken from the optimal decision trees of \citet{bertsimasOptimalClassificationTrees2017},
and the remaining constraints in purple (\ref{eq:Qltsum} -- \ref{eq:accequal2}) and \eqref{eq:constrs}, together with a different objective function \eqref{eq:objective}, are parts of our extensions. We use $[n]$ notation to represent the set of integers $\{1, 2, 3, \ldots, n\}$. An overview table of the variables and parameters is in the Appendix (Table \ref{tab:formdesc}).
}
\label{eq:completeFormulation}
\end{figure} 

\subsection{Base model}
As in the original optimal decision trees \cite{bertsimasOptimalClassificationTrees2017}, we have $n$ samples with $p$ features each. Every point has one of $K$ classes, which is represented in the formulation by a binary matrix $Y$ such that $Y_{ik} = 1$ $\iff$ $y_i = k$. All tree nodes are split into two disjoint sets, $\mathcal{T}_B$ and $\mathcal{T}_L$, which are sets of branching nodes and leaf nodes, respectively.  
Variable $\boldsymbol{a}_{t}$ is a binary vector of dimension $p$ that selects a feature $j$ to be used for decisions in node $t$. It holds that $a_{jt} = 1 \iff j$ is the selected feature in node $t$. $b_t$ is then the value of the threshold. We assume all data are normalized to the $[0, 1]$ range. 

Equations (\ref{eq:orig1}--\ref{eq:origX}) capture the original model of \citet{bertsimasOptimalClassificationTrees2017}, wherein:
\begin{itemize}
\item Binary variable $c_{kt}$ is equal to 1 if and only if leaf node $t$ predicts class $k$ to data. 
\item Binary variable $l_t$ is equal to $1$ if and only if there is any point classified by the leaf node $t$. 
\item Binary variable $z_{it}$ is equal to $1$ if and only if point $x_i$ is classified by leaf node $t$. 
\end{itemize}
The only modification to the original formulation is the omission of a binary variable $d_t$ that indicated whether a certain branching node is used. This introduced a flaw in the original formulation \cite{bertsimasOptimalClassificationTrees2017}, which led to invalid trees, so we decided against using it. We assume it to always be $1$ instead. To prune redundancies, we introduce a process of tree reduction described in Section \ref{sec:reduction}. 

Equations \eqref{eq:decr} and \eqref{eq:decl} implement the split of samples to leaf node $t$ using disjoint sets $A_R(t)$ and $A_L(t)$, containing nodes to which the leaf $t$ is on the right or on the left, respectively. Since we cannot use strict inequality, we use $\boldsymbol{\epsilon}$, a $p$-dimension vector of the smallest increments between two distinct consecutive values in every feature space \cite{bertsimasOptimalClassificationTrees2017}:
{
\footnotesize
$$\epsilon_j = \min\left\{x_j^{(i+1)} - x_j^{(i)} \Big\rvert x_j^{(i+1)} \ne x_j^{(i)}, \forall i \in \{1, \ldots ,n - 1\}\right\}$$
}
where $x_j^{(i)}$ is the $i$-th largest value in the $j$-th feature, $\epsilon_{\max}$ is the highest value of $\epsilon_j$ and serves as a tight big-M bound.

Finally, Equation \eqref{eq:minN} bounds the number of points ($N_{\min}$) in a single leaf from below. 

\subsection{MIO extensions}
In the original optimal decision trees \cite{bertsimasOptimalClassificationTrees2017}, the objective is to minimize total misclassification error. Instead, we wish to maximize the leaf accuracy. 
Because a single sample usually contributes differently to accuracy at different leaves, 
we need to introduce multiple new variables to track the accuracy in each leaf: 
\begin{itemize}
\item variable $s_{it}$ represents the potential accuracy that sample $\boldsymbol{x}_i$ has in leaf $t$. It takes values in the range $[0, 1]$ and must sum to $1$ when summing across all samples assigned to leaf $t$. This is ensured by setting the value to $0$ for all points that are not assigned to the leaf $t$ in constraint \eqref{eq:amount0}. The sum of $1$ is enforced in constraint \eqref{eq:sum1} for non-empty leaves. 
Empty leaves do not have non-zero $s_{it}$ values for any $i$ and thus could not sum to 1. 
\item reference accuracy variable $r_t$ serves as a common variable to which all accuracy contributions are equal. This is, of course, required only for points assigned to the leaf $t$. This is enforced in \eqref{eq:acref1} and \eqref{eq:acref2}. 
\item variable $S_{it}$ represents the true assignment of accuracy given by the sample. That is achieved by setting it to $0$ for misclassified points using constraint \eqref{eq:misclas0} and by setting it equal to $s_{it}$ otherwise by constraints \eqref{eq:accequal1} and \eqref{eq:accequal2}.
\item variable $Q$ is our objective and represents the \emph{leaf accuracy} of the tree. Equivalent to $A_L(T)$ defined in Equation \eqref{eq:leaf_acc}, it is the lowest achieved accuracy across all non-empty leaves as per constraint \eqref{eq:Qltsum}. For empty leaves, this constraint will be trivially satisfied since $Q$ cannot take a value higher than $1$ anyway.  
\end{itemize}

\paragraph{Tree reduction}
\label{sec:reduction}

After the optimized tree is recovered from the formulation, empty leaves are pruned to obtain the resulting unbalanced tree. Furthermore, to account for suboptimal solutions obtained when the solver is run with a strict time limit, each pair of sibling leaves classified in the same class is merged. This is performed recursively until no further action can be performed.
This leads to no loss in model accuracy and oftentimes leads to an improvement in leaf accuracy, given the fact that we consider the minimum over the leaves, which cannot decrease by combining two leaves with the same majority class into one.

\paragraph{Tree extension}
Optionally, we can extend the leaves with further models 
to improve the full model accuracy (hybrid-tree accuracy).
In experiments, we used XGBoost as the extending model 
since it was the best-performing model on the used benchmark \cite{grinsztajnWhyTreebasedModels2022}. We trained a separate model for each leaf of the shallow tree after the aforementioned reduction. The hyperparameters of the models were tuned using 50 iterations of a Bayesian hyperparameter search with 3-fold cross-validation in each leaf. In experiments, we reduce and extend trees generated by other methods (OCT, CART) in the same way.

\section{Numerical results}
\label{sec:results}

We have implemented the method in Python, and 
all code and results are provided in the Supplementary material. 
We will release them under an open-source license to GitHub once the paper has been accepted. 
The hyperparameters have been chosen as follows:
\begin{itemize}
\item The shallow trees have been trained using the formulation in Figure \ref{eq:completeFormulation} with depth limited to four 
since that is a reasonable threshold for interpretability (e.g., printability on an A4 page, similar to Figure~\ref{fig:example}) and for not diluting the dataset to small parts that would impede the ability to train the extending models. 
\item To further support this, we set the minimal amount of points in a leaf ($N_{\min}$) to 50.
\item MIPFocus and Heuristics hyperparameters were set to 1 and 0.8, respectively, to focus on finding feasible solutions in the search since that leads to the fastest improvements of the solution. However, our experiments in Appendix \ref{sec:mip_default} show that default MIO solver hyperparameters perform similarly.  
\end{itemize}

\begin{figure*}[t]
  \centering
  \includegraphics[width=\textwidth]{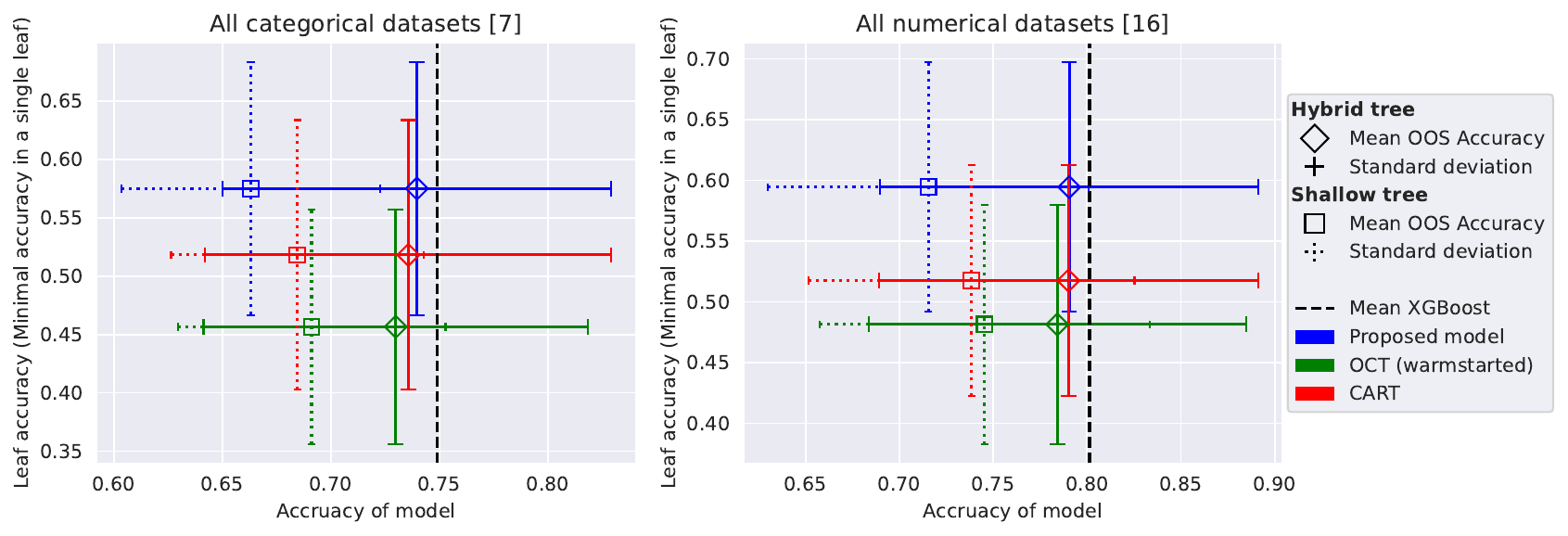}
  \caption{Results on out-of-sample data. The plot shows a significant increase in leaf accuracy when using our method, significantly improving the validity of the explanations provided. It also shows an increase in model accuracy when extending the models with XGBoost models in leaves. The results of the OCT model serve to compare to the model we built upon.}
    \label{fig:agg}
\end{figure*}


We performed our experiments on the benchmark of \citeauthor{grinsztajnWhyTreebasedModels2022}, which contains datasets for both regression and classification. 
The benchmark consists of medium-sized real datasets of tabular data. Tree-based models are the best performing on these datasets \citep{grinsztajnWhyTreebasedModels2022}, making the datasets fitting for our purpose. 
Since our implementation considers only classification,
we consider only classification datasets. 
\citet{grinsztajnWhyTreebasedModels2022} divide the datasets into numerical datasets and datasets with some categorical features. We follow this distinction and present results on both kinds of datasets separately.
We also follow the suggestion of \citeauthor{grinsztajnWhyTreebasedModels2022} to perform 
10 different train-test splits with
at most 10,000 data points or 80\% of total data points (whichever is lower) for training across all datasets.
That is, each model has been trained on each dataset 10 times, with different seeds for data splits. 
The training used 80\% of all data points or 10,000 data points, whichever is lower,
while the remaining 20\% of the dataset has been used as the test set for evaluating the model accuracy and 
leaf accuracy $A_L(T)$---see Equation \eqref{eq:leaf_acc}. All MIO formulations of our shallow tree were warmstarted using a CART solution trained on the same data with default scikit-learn parameters, except for maximal depth and a minimal number of samples in a leaf, which were set to 4 and 50, respectively. 

We performed all experiments on an internal cluster with sufficient amounts of memory. Each run of the MIO solver has been limited to 8 hours on 8 cores of AMD Epyc 7543, totaling 64 core-hours per split of a dataset. The extension part takes, on average, around 1 additional 3 core-hours per split. This totals around 15,500 core-hours for the entire classification part of the tabular benchmark
and one configuration of hyperparameters. 
Training each dataset requires between 15 and 95 GB of working memory; 
details are provided in the Appendix (Figure \ref{fig:memory}). 
In this setting, the Gurobi solver closes the MIP Gap to around 60\% on average. Further discussion of MIP Gaps is in Appendix \ref{sec:mip_gaps}. Still, the proposed model outperforms CART even after just 1 hour of computation; see Figure \ref{fig:performance_1hour} in Appendix.


We compare our method of training classification trees to CART, as it is by far the most common. All experiments used the scikit-learn implementation of CART, also utilizing the option of cost complexity pruning. The hyperparameters for CART were optimized using Bayesian hyperparameter optimization for 100 iterations using 5-fold cross-validation. Hyperparameter search space was notably constrained only by fixing a maximal depth to 4 and a minimal number of samples in leaves to 50, ensuring comparability to our shallow trees. In comparison to unconstrained depth CART and CART with optimized lower bound on the number of samples in a leaf, our model interestingly fared even better. See Appendix \ref{sec:unlim_cart} for details.
The entire optimization of CART with the extensions of the leaves took around 500 core-hours for the entire benchmark.

The XGBoost results are taken from the authors of the paper introducing the benchmark, which suggests 20,000 core-hours have been spent producing these. \cite{grinsztajnWhyTreebasedModels2022}

Figure \ref{fig:agg} shows the average performance (model accuracy and leaf accuracy) over categorical and numerical datasets. We include the comparison to optimal classification trees (OCT) \cite{bertsimasOptimalClassificationTrees2017} since it is the formulation we built on. The OCT model is warmstarted the same way as the proposed model and has the worst performance. The proposed model improves the leaf accuracy by 7.09 percentage points on average compared to CART.


\begin{table}[t]
    \centering
    \caption{Improvements in mean accuracy on datasets between our model and comparable models. Data is computed by subtracting the mean accuracy of CART or XGBoost, respectively, on each dataset from the mean accuracy of our model. In the first two rows, we compare the leaf accuracy of our shallow model to CART. In the middle two rows, we compare the hybrid-tree accuracy of the extended trees with CART trees extended in the same way. In the last two rows, we compare our extended hybrid-tree model to the mean XGBoost model trained on the same dataset.}
    \label{tab:acc_differences}
    \resizebox{\textwidth}{!}{\begin{tabular}{lrrrrr}
            \toprule
         & \textbf{Baseline} &\textbf{Data type} & \textbf{Minimum} & \textbf{Mean ($\pm$ std)} & \textbf{Maximum} \\ 
         \midrule
         \midrule
         \textbf{Leaf Accuracy} & CART & categorical & $-0.0142$ & $0.0569 \pm0.0533$ & $0.1206$ \\
         & & numerical & $-0.0061$ & $0.0770 \pm0.0556$ & $0.1841$ \\
         \midrule
         \textbf{Hybrid-tree Accuracy} & CART & categorical & $-0.0078$ & $0.0040 \pm0.0071$ & $0.0147$ \\
         & & numerical & $-0.0244$ & $0.0004 \pm0.0082$ & $0.0087$ \\   
         \midrule
         \textbf{Hybrid-tree Accuracy} &  XGBoost & categorical & $-0.0228$ & $-0.0095 \pm0.0064$ & $-0.0036$ \\
         & & numerical & $-0.0276$ & $-0.0108 \pm0.0076$ & $0.0005$ \\
         \bottomrule
    \end{tabular}}
    
\end{table}

Table \ref{tab:acc_differences} quantifies the differences numerically. Our partially interpretable model has worse accuracy by about 1 percentage point on average when compared to the uninterpretable, best-performing state-of-the-art model (XGBoost). Our approach improves the accuracy of hybrid-tree models built on CART trees. But more importantly, it improves the leaf accuracy.


\paragraph{Statistical significance}
\citet{demsarStatisticalComparisonsClassifiers2006} summarizes statistical tests used for the comparison of algorithms on multiple datasets. Compared to CART, the proposed method has better leaf accuracy and better hybrid-tree accuracy on a substantial majority of datasets. Using the basic sign test \cite{demsarStatisticalComparisonsClassifiers2006}, both results are statistically significant with $p < 0.05$. 
Using the Wilcoxon signed-rank test, we reject the null hypothesis that CART performs better than the proposed method with high confidence ($\alpha \le 0.01$ for leaf accuracy and $\alpha \le 0.05$ for hybrid-tree accuracy).
For further results, refer to Appendix \ref{sec:detailed}.

\section{Conclusions and Limitations}

We have identified an important problem of the \emph{fairness} and \emph{validity} of a tree as an explanation and have shown that contemporary tree-based models do leave room for improvement in terms of fairness.
Our approach offers multiple benefits. 

First, it ensures better validity of every explanation provided, improving the leaf accuracy by around 7 percentage points on average across the benchmark of tabular datasets \cite{grinsztajnWhyTreebasedModels2022}. It improves the fairness of single explanations by narrowing the difference between the accuracies of the leaf, i.e., explanation.

Second, if extended, the hybrid-tree accuracy with black-box models extending the leaves improves over the accuracy of shallow trees constructed using integer optimization as well as hybrid trees, where the shallow tree is obtained using CART. 

Finally, it is easy to extend to further constraints, such as shape constraints, in the top tree. 
Overall, we hope that the proposed approach may lead to improving the validity and fairness of decision trees as explanations. 

\paragraph{Limitations}

The shallow tree can explain a significant portion of the data while ensuring the global explanation provided is fair and valid for all paths from the root, i.e., for all subgroups. This limitation is common in partially interpretable models.

Similar to most partially interpretable models, our hybrid-tree model aims to strike a balance between global explainability and model accuracy. The extending black-box models limit the use of the shallow tree as an explanation, especially in cases when the extending model changes the decision of the shallow tree. 
In our experiments, 
the shallow tree agrees with the extending models in 62.3\% 
of the cases on average across all the datasets tested. Our approach thus allows the choice between a fair and explainable shallow tree with lower accuracy or a highly accurate model that provides a simple explanation for a majority of the cases. This ``agreement rate'' for the proposed method is comparable to the rate for CART trees (63.5\%), despite shallow CART trees' higher accuracy.
Explaining decisions where there is no agreement requires the use of other explanation methods. 

The proposed approach shares some of the scalability limitations of the original optimal decision trees \cite{bertsimasOptimalClassificationTrees2017}. 
Notably, the algorithms we utilize for solving mixed-integer optimization problems scale exponentially in the number of decision variables.
Having said that, depth-4 trees suffice to match state-of-the-art methods in terms of accuracy
when additional tree-based models extend from the leaves,
which makes exponential time algorithms sufficiently fast in practice. 
Furthermore, all recently proposed methods \citep[e.g.,][]{pmlr-v119-vidal20a,demirovic2022murtree,van2022fair,hua2022scalable,Mazumder2022} improving the scalability of optimal decision trees can be applied, in principle.  

\subsection*{Acknowledgment}

This work has received funding from the European Union’s Horizon Europe research and
innovation program under grant agreement No. 101070568.


\bibliographystyle{plainnat}
\bibliography{NeurIPS2024/explainability}

\begin{thebibliography}{64}
\providecommand{\natexlab}[1]{#1}
\providecommand{\url}[1]{\texttt{#1}}
\expandafter\ifx\csname urlstyle\endcsname\relax
  \providecommand{\doi}[1]{doi: #1}\else
  \providecommand{\doi}{doi: \begingroup \urlstyle{rm}\Url}\fi

\bibitem[Aghaei et~al.(2020)Aghaei, Gomez, and Vayanos]{aghaei2020learning}
Sina Aghaei, Andres Gomez, and Phebe Vayanos.
\newblock Learning optimal classification trees: Strong max-flow formulations.
\newblock \emph{arXiv preprint arXiv:2002.09142}, 2020.

\bibitem[Arrieta et~al.(2020)Arrieta, D{\'\i}az-Rodr{\'\i}guez, Del~Ser,
  Bennetot, Tabik, Barbado, Garc{\'\i}a, Gil-L{\'o}pez, Molina, Benjamins,
  et~al.]{arrieta2020explainable}
Alejandro~Barredo Arrieta, Natalia D{\'\i}az-Rodr{\'\i}guez, Javier Del~Ser,
  Adrien Bennetot, Siham Tabik, Alberto Barbado, Salvador Garc{\'\i}a, Sergio
  Gil-L{\'o}pez, Daniel Molina, Richard Benjamins, et~al.
\newblock Explainable artificial intelligence (xai): Concepts, taxonomies,
  opportunities and challenges toward responsible ai.
\newblock \emph{Information fusion}, 58:\penalty0 82--115, 2020.

\bibitem[Athey and Imbens(2019)]{athey2019machine}
Susan Athey and Guido~W Imbens.
\newblock Machine learning methods that economists should know about.
\newblock \emph{Annual Review of Economics}, 11:\penalty0 685--725, 2019.

\bibitem[Bastani et~al.(2017)Bastani, Kim, and
  Bastani]{bastani2017interpreting}
Osbert Bastani, Carolyn Kim, and Hamsa Bastani.
\newblock Interpreting blackbox models via model extraction.
\newblock \emph{arXiv preprint arXiv:1705.08504}, 2017.

\bibitem[Bertsimas and Dunn(2017)]{bertsimasOptimalClassificationTrees2017}
Dimitris Bertsimas and Jack Dunn.
\newblock Optimal classification trees.
\newblock \emph{Machine Learning}, 106\penalty0 (7):\penalty0 1039--1082, July
  2017.
\newblock ISSN 1573-0565.
\newblock \doi{10.1007/s10994-017-5633-9}.

\bibitem[Bessiere et~al.(2009)Bessiere, Hebrard, and O'Sullivan]{Bessiere2009}
Christian Bessiere, Emmanuel Hebrard, and Barry O'Sullivan.
\newblock Minimising decision tree size as combinatorial optimisation.
\newblock In \emph{Proceedings of the 15th International Conference on
  Principles and Practice of Constraint Programming}, CP'09, page 173–187,
  Berlin, Heidelberg, 2009. Springer-Verlag.
\newblock ISBN 3642042430.

\bibitem[Bracke et~al.(2019)Bracke, Datta, Jung, and
  Sen]{BritishDiscussionPaper}
Philippe Bracke, Anupam Datta, Carsten Jung, and Shayak Sen.
\newblock Machine learning explainability in finance: an application to default
  risk analysis.
\newblock Staff Working Paper No. 816 of the Bank of England,
  \url{https://www.bankofengland.co.uk/working-paper/2019/}, 2019.
\newblock Accessed: 2023-04-30.

\bibitem[Breiman et~al.(1984)Breiman, Friedman, Stone, and
  Olshen]{breiman1984classification}
L.~Breiman, J.~Friedman, C.J. Stone, and R.A. Olshen.
\newblock \emph{Classification and Regression Trees}.
\newblock Taylor \& Francis, 1984.
\newblock ISBN 9780412048418.

\bibitem[Breiman(1998)]{breiman1998arcing}
Leo Breiman.
\newblock Arcing classifier (with discussion and a rejoinder by the author).
\newblock \emph{The annals of statistics}, 26\penalty0 (3):\penalty0 801--849,
  1998.

\bibitem[Breiman(2001)]{breiman2001random}
Leo Breiman.
\newblock Random forests.
\newblock \emph{Machine learning}, 45:\penalty0 5--32, 2001.

\bibitem[Brennan and Dieterich(2018)]{Brennan2018}
Tim Brennan and William Dieterich.
\newblock \emph{Correctional Offender Management Profiles for Alternative
  Sanctions (COMPAS)}, chapter~3, pages 49--75.
\newblock John Wiley \& Sons, Ltd, 2018.
\newblock ISBN 9781119184256.
\newblock \doi{https://doi.org/10.1002/9781119184256.ch3}.
\newblock URL
  \url{https://onlinelibrary.wiley.com/doi/abs/10.1002/9781119184256.ch3}.

\bibitem[Brennan et~al.(2009)Brennan, Dieterich, and
  Ehret]{brennan2009evaluating}
Tim Brennan, William Dieterich, and Beate Ehret.
\newblock Evaluating the predictive validity of the compas risk and needs
  assessment system.
\newblock \emph{Criminal Justice and behavior}, 36\penalty0 (1):\penalty0
  21--40, 2009.

\bibitem[Burkart and Huber(2021)]{burkart2021survey}
Nadia Burkart and Marco~F Huber.
\newblock A survey on the explainability of supervised machine learning.
\newblock \emph{Journal of Artificial Intelligence Research}, 70:\penalty0
  245--317, 2021.

\bibitem[Carreira-Perpin{\'a}n and Tavallali(2018)]{carreira2018alternating}
Miguel~A Carreira-Perpin{\'a}n and Pooya Tavallali.
\newblock Alternating optimization of decision trees, with application to
  learning sparse oblique trees.
\newblock \emph{Advances in neural information processing systems}, 31, 2018.

\bibitem[Carrizosa et~al.(2021)Carrizosa, Molero-Rio, and
  Romero~Morales]{carrizosa2021mathematical}
Emilio Carrizosa, Cristina Molero-Rio, and Dolores Romero~Morales.
\newblock Mathematical optimization in classification and regression trees.
\newblock \emph{Top}, 29\penalty0 (1):\penalty0 5--33, 2021.

\bibitem[Chen and Guestrin(2016)]{Tianqi2016}
Tianqi Chen and Carlos Guestrin.
\newblock Xgboost: A scalable tree boosting system.
\newblock In \emph{Proceedings of the 22nd ACM SIGKDD International Conference
  on Knowledge Discovery and Data Mining}, KDD '16, page 785–794, New York,
  NY, USA, 2016. Association for Computing Machinery.
\newblock ISBN 9781450342322.
\newblock \doi{10.1145/2939672.2939785}.

\bibitem[{Consumer Financial Protection Bureau}(2022)]{Circular}
{Consumer Financial Protection Bureau}.
\newblock Consumer financial protection circular 2022-03: Adverse action
  notification requirements in connection with credit decisions based on
  complex algorithms.
\newblock \url{https://www.consumerfinance.gov/compliance/circulars/}, 2022.
\newblock Accessed: 2023-04-30.

\bibitem[Courtland(2018)]{courtland2018bias}
Rachel Courtland.
\newblock The bias detectives.
\newblock \emph{Nature}, 558\penalty0 (7710):\penalty0 357--360, 2018.

\bibitem[Dash et~al.(2018)Dash, Gunluk, and Wei]{dash2018boolean}
Sanjeeb Dash, Oktay Gunluk, and Dennis Wei.
\newblock Boolean decision rules via column generation.
\newblock \emph{Advances in neural information processing systems}, 31, 2018.

\bibitem[Demirovi{\'c} et~al.(2022)Demirovi{\'c}, Lukina, Hebrard, Chan,
  Bailey, Leckie, Ramamohanarao, and Stuckey]{demirovic2022murtree}
Emir Demirovi{\'c}, Anna Lukina, Emmanuel Hebrard, Jeffrey Chan, James Bailey,
  Christopher Leckie, Kotagiri Ramamohanarao, and Peter~J Stuckey.
\newblock Murtree: Optimal decision trees via dynamic programming and search.
\newblock \emph{The Journal of Machine Learning Research}, 23\penalty0
  (1):\penalty0 1169--1215, 2022.

\bibitem[Dem{\v s}ar(2006)]{demsarStatisticalComparisonsClassifiers2006}
Janez Dem{\v s}ar.
\newblock Statistical {{Comparisons}} of {{Classifiers}} over {{Multiple Data
  Sets}}.
\newblock \emph{The Journal of Machine Learning Research}, 7:\penalty0 1--30,
  December 2006.
\newblock ISSN 1532-4435.

\bibitem[Dupont et~al.(2020)Dupont, Fliche, and Yang]{FrenchDiscussionPaper}
Laurent Dupont, Olivier Fliche, and Su~Yang.
\newblock Governance of artificial intelligence in finance.
\newblock Discussion papers of Autorité de Contrôle Prudentiel et de
  Résolution,
  \url{https://acpr.banque-france.fr/en/governance-artificial-intelligence-finance},
  2020.
\newblock Accessed: 2023-04-30.

\bibitem[{Equal Credit Opportunity Act} ({ECOA})()]{ECOA}
{Equal Credit Opportunity Act} ({ECOA}).
\newblock {Equal Credit Opportunity Act} ({ECOA}).
\newblock
  \url{https://www.law.cornell.edu/uscode/text/15/chapter-41/subchapter-IV},
  1974.
\newblock Title 15 of the United States Code, Chapter 41, Subchapter IV,
  paragraph 1691 and following.

\bibitem[{European Commission}(2016{\natexlab{a}})]{CreditDir}
{European Commission}.
\newblock Directive {2013/36/EU of the European Parliament and of the Council}
  of 26 june 2013 on access to the activity of credit institutions and the
  prudential supervision of credit institutions and investment firms, amending
  {Directive 2002/87/EC} and repealing {Directives 2006/48/EC} and 2006/49/ec.,
  2016{\natexlab{a}}.
\newblock URL
  \url{https://eur-lex.europa.eu/legal-content/EN/TXT/?uri=celex%3A32013L0036}.
\newblock Accessed: 2023-04-30.

\bibitem[{European Commission}(2016{\natexlab{b}})]{GDPR}
{European Commission}.
\newblock Regulation ({EU}) 2016/679 of the {European} {Parliament} and of the
  {Council} of 27 {April} 2016 on the protection of natural persons with regard
  to the processing of personal data and on the free movement of such data, and
  repealing {Directive} 95/46/{EC} ({General} {Data} {Protection}
  {Regulation})., 2016{\natexlab{b}}.
\newblock URL \url{https://eur-lex.europa.eu/eli/reg/2016/679/oj}.
\newblock Accessed: 2023-04-30.

\bibitem[Feldman(2000)]{feldman2000minimization}
Jacob Feldman.
\newblock Minimization of boolean complexity in human concept learning.
\newblock \emph{Nature}, 407\penalty0 (6804):\penalty0 630--633, 2000.

\bibitem[Friedman(2001)]{Friedman2001}
Jerome~H. Friedman.
\newblock {Greedy function approximation: A gradient boosting machine.}
\newblock \emph{The Annals of Statistics}, 29\penalty0 (5):\penalty0 1189 --
  1232, 2001.
\newblock \doi{10.1214/aos/1013203451}.
\newblock URL \url{https://doi.org/10.1214/aos/1013203451}.

\bibitem[Frost et~al.(2024)Frost, Lipton, Mansour, and
  Moshkovitz]{frostPartiallyInterpretableModels2024}
Nave Frost, Zachary Lipton, Yishay Mansour, and Michal Moshkovitz.
\newblock Partially {{Interpretable Models}} with {{Guarantees}} on
  {{Coverage}} and {{Accuracy}}.
\newblock In \emph{Proceedings of {{The}} 35th {{International Conference}} on
  {{Algorithmic Learning Theory}}}, pages 590--613. PMLR, March 2024.

\bibitem[Gorishniy et~al.(2021)Gorishniy, Rubachev, Khrulkov, and
  Babenko]{NEURIPS2021_9d86d83f}
Yury Gorishniy, Ivan Rubachev, Valentin Khrulkov, and Artem Babenko.
\newblock Revisiting deep learning models for tabular data.
\newblock In M.~Ranzato, A.~Beygelzimer, Y.~Dauphin, P.S. Liang, and J.~Wortman
  Vaughan, editors, \emph{Advances in Neural Information Processing Systems},
  volume~34, pages 18932--18943. Curran Associates, Inc., 2021.
\newblock URL
  \url{https://proceedings.neurips.cc/paper_files/paper/2021/file/9d86d83f925f2149e9edb0ac3b49229c-Paper.pdf}.

\bibitem[Grinsztajn et~al.(2022)Grinsztajn, Oyallon, and
  Varoquaux]{grinsztajnWhyTreebasedModels2022}
L{\'e}o Grinsztajn, Edouard Oyallon, and Ga{\"e}l Varoquaux.
\newblock Why do tree-based models still outperform deep learning on typical
  tabular data?
\newblock In \emph{Thirty-sixth Conference on Neural Information Processing
  Systems Datasets and Benchmarks Track}, 2022.
\newblock URL \url{https://openreview.net/forum?id=Fp7__phQszn}.

\bibitem[G{\"u}nl{\"u}k et~al.(2021)G{\"u}nl{\"u}k, Kalagnanam, Li, Menickelly,
  and Scheinberg]{gunluk2021optimal}
Oktay G{\"u}nl{\"u}k, Jayant Kalagnanam, Minhan Li, Matt Menickelly, and Katya
  Scheinberg.
\newblock Optimal decision trees for categorical data via integer programming.
\newblock \emph{Journal of Global Optimization}, 81:\penalty0 233--260, 2021.
\newblock First appeared in a pre-print form in 2016 as arXiv:1612.03225.

\bibitem[Gunnarsson et~al.(2021)Gunnarsson, Vanden~Broucke, Baesens,
  {\'O}skarsd{\'o}ttir, and Lemahieu]{gunnarsson2021deep}
Bj{\"o}rn~Rafn Gunnarsson, Seppe Vanden~Broucke, Bart Baesens, Mar{\'\i}a
  {\'O}skarsd{\'o}ttir, and Wilfried Lemahieu.
\newblock Deep learning for credit scoring: Do or don’t?
\newblock \emph{European Journal of Operational Research}, 295\penalty0
  (1):\penalty0 292--305, 2021.

\bibitem[Hu et~al.(2019)Hu, Rudin, and Seltzer]{hu2019optimal}
Xiyang Hu, Cynthia Rudin, and Margo Seltzer.
\newblock Optimal sparse decision trees.
\newblock \emph{Advances in Neural Information Processing Systems}, 32, 2019.

\bibitem[Hua et~al.(2022)Hua, Ren, and Cao]{hua2022scalable}
Kaixun Hua, Jiayang Ren, and Yankai Cao.
\newblock A scalable deterministic global optimization algorithm for training
  optimal decision tree.
\newblock \emph{Advances in Neural Information Processing Systems},
  35:\penalty0 8347--8359, 2022.

\bibitem[Ke et~al.(2017)Ke, Meng, Finley, Wang, Chen, Ma, Ye, and
  Liu]{NIPS2017_6449f44a}
Guolin Ke, Qi~Meng, Thomas Finley, Taifeng Wang, Wei Chen, Weidong Ma, Qiwei
  Ye, and Tie-Yan Liu.
\newblock Lightgbm: A highly efficient gradient boosting decision tree.
\newblock In I.~Guyon, U.~Von Luxburg, S.~Bengio, H.~Wallach, R.~Fergus,
  S.~Vishwanathan, and R.~Garnett, editors, \emph{Advances in Neural
  Information Processing Systems}, volume~30. Curran Associates, Inc., 2017.

\bibitem[Laurent and Rivest(1976)]{laurent1976constructing}
Hyafil Laurent and Ronald~L Rivest.
\newblock Constructing optimal binary decision trees is np-complete.
\newblock \emph{Information processing letters}, 5\penalty0 (1):\penalty0
  15--17, 1976.

\bibitem[Le and Clarke(2020)]{lePraisePartiallyInterpretable2020}
Tri Le and Bertrand Clarke.
\newblock In praise of partially interpretable predictors.
\newblock \emph{Statistical Analysis and Data Mining: The ASA Data Science
  Journal}, 13\penalty0 (2):\penalty0 113--133, 2020.
\newblock ISSN 1932-1872.
\newblock \doi{10.1002/sam.11450}.

\bibitem[Lessmann et~al.(2015)Lessmann, Baesens, Seow, and
  Thomas]{lessmann2015benchmarking}
Stefan Lessmann, Bart Baesens, Hsin-Vonn Seow, and Lyn~C Thomas.
\newblock Benchmarking state-of-the-art classification algorithms for credit
  scoring: An update of research.
\newblock \emph{European Journal of Operational Research}, 247\penalty0
  (1):\penalty0 124--136, 2015.

\bibitem[Lin et~al.(2020)Lin, Zhong, Hu, Rudin, and
  Seltzer]{lin2020generalized}
Jimmy Lin, Chudi Zhong, Diane Hu, Cynthia Rudin, and Margo Seltzer.
\newblock Generalized and scalable optimal sparse decision trees.
\newblock In \emph{International Conference on Machine Learning}, pages
  6150--6160. PMLR, 2020.

\bibitem[London(2019)]{london2019artificial}
Alex~John London.
\newblock Artificial intelligence and black-box medical decisions: accuracy
  versus explainability.
\newblock \emph{Hastings Center Report}, 49\penalty0 (1):\penalty0 15--21,
  2019.

\bibitem[Mason et~al.(1999)Mason, Baxter, Bartlett, and
  Frean]{mason1999boosting}
Llew Mason, Jonathan Baxter, Peter Bartlett, and Marcus Frean.
\newblock Boosting algorithms as gradient descent.
\newblock \emph{Advances in neural information processing systems}, 12, 1999.

\bibitem[Mays(1995)]{mays1995handbook}
Elizabeth Mays.
\newblock \emph{Handbook of credit scoring}.
\newblock Global Professional Publishing, 1995.

\bibitem[Mazumder et~al.(2022)Mazumder, Meng, and Wang]{Mazumder2022}
Rahul Mazumder, Xiang Meng, and Haoyue Wang.
\newblock Quant-{B}n{B}: A scalable branch-and-bound method for optimal
  decision trees with continuous features.
\newblock In Kamalika Chaudhuri, Stefanie Jegelka, Le~Song, Csaba Szepesvari,
  Gang Niu, and Sivan Sabato, editors, \emph{Proceedings of the 39th
  International Conference on Machine Learning}, volume 162 of
  \emph{Proceedings of Machine Learning Research}, pages 15255--15277. PMLR,
  17--23 Jul 2022.
\newblock URL \url{https://proceedings.mlr.press/v162/mazumder22a.html}.

\bibitem[Nanfack et~al.(2022)Nanfack, Temple, and Fr\'{e}nay]{Nanfack2022}
G\'{e}raldin Nanfack, Paul Temple, and Beno\^{\i}t Fr\'{e}nay.
\newblock Constraint enforcement on decision trees: A survey.
\newblock \emph{ACM Comput. Surv.}, 54\penalty0 (10s), sep 2022.
\newblock ISSN 0360-0300.
\newblock \doi{10.1145/3506734}.
\newblock URL \url{https://doi.org/10.1145/3506734}.

\bibitem[Narodytska et~al.(2018)Narodytska, Ignatiev, Pereira, and
  Marques-Silva]{Narodytska2018}
Nina Narodytska, Alexey Ignatiev, Filipe Pereira, and Joao Marques-Silva.
\newblock Learning optimal decision trees with sat.
\newblock In \emph{Proceedings of the 27th International Joint Conference on
  Artificial Intelligence}, IJCAI'18, page 1362–1368. AAAI Press, 2018.
\newblock ISBN 9780999241127.

\bibitem[Payne and Meisel(1977)]{Payne1977AnAF}
Harold~J. Payne and William~S. Meisel.
\newblock An algorithm for constructing optimal binary decision trees.
\newblock \emph{IEEE Transactions on Computers}, C-26:\penalty0 905--916, 1977.

\bibitem[Prokhorenkova et~al.(2018)Prokhorenkova, Gusev, Vorobev, Dorogush, and
  Gulin]{prokhorenkova2018catboost}
Liudmila Prokhorenkova, Gleb Gusev, Aleksandr Vorobev, Anna~Veronika Dorogush,
  and Andrey Gulin.
\newblock Catboost: unbiased boosting with categorical features.
\newblock \emph{Advances in neural information processing systems}, 31, 2018.

\bibitem[Rakha et~al.(2014)Rakha, Soria, Green, Lemetre, Powe, Nolan,
  Garibaldi, Ball, and Ellis]{rakha2014nottingham}
EA~Rakha, Daniele Soria, Andrew~R Green, Christophe Lemetre, Desmond~G Powe,
  Christopher~C Nolan, Jonathan~M Garibaldi, Graham Ball, and Ian~O Ellis.
\newblock Nottingham prognostic index plus (npi+): a modern clinical decision
  making tool in breast cancer.
\newblock \emph{British journal of cancer}, 110\penalty0 (7):\penalty0
  1688--1697, 2014.

\bibitem[Rudin(2019)]{rudin2019stop}
Cynthia Rudin.
\newblock Stop explaining black box machine learning models for high stakes
  decisions and use interpretable models instead.
\newblock \emph{Nature Machine Intelligence}, 1\penalty0 (5):\penalty0
  206--215, 2019.

\bibitem[Rudin et~al.(2022)Rudin, Chen, Chen, Huang, Semenova, and
  Zhong]{rudin2022interpretable}
Cynthia Rudin, Chaofan Chen, Zhi Chen, Haiyang Huang, Lesia Semenova, and Chudi
  Zhong.
\newblock Interpretable machine learning: Fundamental principles and 10 grand
  challenges.
\newblock \emph{Statistic Surveys}, 16:\penalty0 1--85, 2022.

\bibitem[Thomas et~al.(2017)Thomas, Crook, and Edelman]{thomas2017credit}
Lyn Thomas, Jonathan Crook, and David Edelman.
\newblock \emph{Credit scoring and its applications}.
\newblock SIAM, 2017.

\bibitem[Tjoa and Guan(2020)]{tjoa2020survey}
Erico Tjoa and Cuntai Guan.
\newblock A survey on explainable artificial intelligence ({XAI}): Toward
  medical {XAI}.
\newblock \emph{IEEE transactions on neural networks and learning systems},
  32\penalty0 (11):\penalty0 4793--4813, 2020.

\bibitem[van~der Linden et~al.(2022)van~der Linden, de~Weerdt, and
  Demirovi{\'c}]{van2022fair}
Jacobus van~der Linden, Mathijs de~Weerdt, and Emir Demirovi{\'c}.
\newblock Fair and optimal decision trees: A dynamic programming approach.
\newblock \emph{Advances in Neural Information Processing Systems},
  35:\penalty0 38899--38911, 2022.

\bibitem[Verwer and Zhang(2019)]{Verwer_Zhang_2019}
Sicco Verwer and Yingqian Zhang.
\newblock Learning optimal classification trees using a binary linear program
  formulation.
\newblock \emph{Proceedings of the AAAI Conference on Artificial Intelligence},
  33\penalty0 (01):\penalty0 1625--1632, Jul. 2019.
\newblock \doi{10.1609/aaai.v33i01.33011624}.
\newblock URL \url{https://ojs.aaai.org/index.php/AAAI/article/view/3978}.

\bibitem[Vidal and Schiffer(2020)]{pmlr-v119-vidal20a}
Thibaut Vidal and Maximilian Schiffer.
\newblock Born-again tree ensembles.
\newblock In Hal~Daumé III and Aarti Singh, editors, \emph{Proceedings of the
  37th International Conference on Machine Learning}, volume 119 of
  \emph{Proceedings of Machine Learning Research}, pages 9743--9753. PMLR,
  13--18 Jul 2020.
\newblock URL \url{https://proceedings.mlr.press/v119/vidal20a.html}.

\bibitem[Wang(2019-06-09/2019-06-15)]{wangGainingFreeLowcost2019}
Tong Wang.
\newblock Gaining free or low-cost interpretability with interpretable partial
  substitute.
\newblock In Kamalika Chaudhuri and Ruslan Salakhutdinov, editors,
  \emph{Proceedings of the 36th International Conference on Machine Learning},
  volume~97 of \emph{Proceedings of Machine Learning Research}, pages
  6505--6514. PMLR, 2019-06-09/2019-06-15.

\bibitem[Wolsey(2021)]{wolseyIntegerProgramming2021}
Laurence~A. Wolsey.
\newblock \emph{Integer Programming}.
\newblock {Wiley}, {Hoboken, NJ}, second edition edition, 2021.
\newblock ISBN 978-1-119-60655-0 978-1-119-60652-9.

\bibitem[Wu et~al.(2008)Wu, Kumar, Ross~Quinlan, Ghosh, Yang, Motoda,
  McLachlan, Ng, Liu, Yu, et~al.]{wu2008top}
Xindong Wu, Vipin Kumar, J~Ross~Quinlan, Joydeep Ghosh, Qiang Yang, Hiroshi
  Motoda, Geoffrey~J McLachlan, Angus Ng, Bing Liu, Philip~S Yu, et~al.
\newblock Top 10 algorithms in data mining.
\newblock \emph{Knowledge and information systems}, 14:\penalty0 1--37, 2008.

\bibitem[Xin et~al.(2022)Xin, Zhong, Chen, Takagi, Seltzer, and
  Rudin]{xin2022exploring}
Rui Xin, Chudi Zhong, Zhi Chen, Takuya Takagi, Margo Seltzer, and Cynthia
  Rudin.
\newblock Exploring the whole rashomon set of sparse decision trees.
\newblock \emph{arXiv preprint arXiv:2209.08040}, 2022.

\bibitem[Zantedeschi et~al.(2021)Zantedeschi, Kusner, and
  Niculae]{zantedeschi2021learning}
Valentina Zantedeschi, Matt Kusner, and Vlad Niculae.
\newblock Learning binary decision trees by argmin differentiation.
\newblock In \emph{International Conference on Machine Learning}, pages
  12298--12309. PMLR, 2021.

\bibitem[Zhang et~al.(2023)Zhang, Xin, Seltzer, and Rudin]{ZhangEtAlOSRT2023}
Rui Zhang, Rui Xin, Margo Seltzer, and Cynthia Rudin.
\newblock Optimal sparse regression trees.
\newblock In \emph{{AAAI} Conference on Artificial Intelligence ({AAAI})},
  2023.

\bibitem[Zhou et~al.(2023)Zhou, Marecek, and Shorten]{Zhou2023}
Quan Zhou, Jakub Marecek, and Robert~N Shorten.
\newblock Fairness in forecasting of observations of linear dynamical systems.
\newblock \emph{Journal of AI Research}, 76:\penalty0 1245--1280, 2023.
\newblock arXiv preprint arXiv:2209.05274.

\bibitem[Zhou and Chen(2002)]{zhou2002hybrid}
Zhi-Hua Zhou and Zhao-Qian Chen.
\newblock Hybrid decision tree.
\newblock \emph{Knowledge-based systems}, 15\penalty0 (8):\penalty0 515--528,
  2002.

\bibitem[Zhu et~al.(2020)Zhu, Murali, Phan, Nguyen, and
  Kalagnanam]{zhu2020scalable}
Haoran Zhu, Pavankumar Murali, Dzung Phan, Lam Nguyen, and Jayant Kalagnanam.
\newblock A scalable mip-based method for learning optimal multivariate
  decision trees.
\newblock \emph{Advances in neural information processing systems},
  33:\penalty0 1771--1781, 2020.

\end{thebibliography}


\appendix

\section{Appendix}

In the Supplementary material, we provide the source code, complete results in \texttt{.csv} files, and a Jupyter notebook with example tests. All will be publicly available once the paper is accepted. Here, we present the results of further tests performed, describe ablation analyses, and provide further details about the results already presented.

\subsection{Datasets}
We used the classification part of the data sets from the mid-sized tabular data put together by \citet{grinsztajnWhyTreebasedModels2022}. The datasets, with their properties, are listed in Table \ref{tab:data}. Training sets contained 80\% of the total amount of samples truncated to at most 10,000 samples. This constraint affects 16 of the 23 total datasets, although some only marginally. The affected datasets have their number of samples in Table \ref{tab:data} in bold. The remaining 20\% of the samples were the testing dataset. 
We used 10 random seeds that determined the train-test splits of each dataset and fixed the randomness in the training process. The seeds were namely integers 0 to 9.

Additionally, datasets are either categorical or numerical. Categorical are those that contain at least one categorical feature. Numerical datasets have no categorical features.
Four numerical datasets are the same as categorical datasets but with their categorical features removed (\texttt{covertype}, \texttt{default-of-credit-card-clients}, \texttt{electricity}, \texttt{eye\_movements}).
Only datasets without missing features and with sufficient complexity are included in the benchmark. For more details on the methodology of dataset selection, we refer to the original paper \citep{grinsztajnWhyTreebasedModels2022}.

\begin{table*}
    \centering
    \caption{Listed classification datasets of the tabular benchmark. Train sets contained 80\% of the total amount of samples truncated to at most 10 000 samples. 16 datasets affected by this have their number of samples in bold.}
    \label{tab:data}
    \begin{tabular}{llll}
    \toprule
    \textbf{categorical datasets} & \# of samples & \# of features & \# of classes \\
    \midrule
    albert                         &        \textbf{58252} &            31 &            2 \\
    compas-two-years               &         4966 &            11 &            2 \\
    covertype                      &       \textbf{423680} &            54 &            2 \\
    default-of-credit-card-clients &        \textbf{13272} &            21 &            2 \\
    electricity                    &        \textbf{38474} &             8 &            2 \\
    eye\_movements                  &         7608 &            23 &            2 \\
    road-safety                    &       \textbf{111762} &            32 &            2 \\
    \bottomrule
    \toprule
    \textbf{numerical datasets} & \# of samples & \# of features & \# of classes \\
    \midrule
    bank-marketing                 &        10578 &             7 &            2 \\
    Bioresponse                    &         3434 &           419 &            2 \\
    california                     &        \textbf{20634} &             8 &            2 \\
    covertype                      &       \textbf{566602} &            32 &            2 \\
    credit                         &        \textbf{16714} &            10 &            2 \\
    default-of-credit-card-clients &        \textbf{13272} &            20 &            2 \\
    Diabetes130US                  &        \textbf{71090} &             7 &            2 \\
    electricity                    &        \textbf{38474} &             7 &            2 \\
    eye\_movements                  &         7608 &            20 &            2 \\
    Higgs                          &       \textbf{940160} &            24 &            2 \\
    heloc                          &        10000 &            22 &            2 \\
    house\_16H                      &        \textbf{13488} &            16 &            2 \\
    jannis                         &        \textbf{57580} &            54 &            2 \\
    MagicTelescope                 &        \textbf{13376} &            10 &            2 \\
    MiniBooNE                      &        \textbf{72998} &            50 &            2 \\
    pol                            &        10082 &            26 &            2 \\
    \bottomrule
    \end{tabular}
\end{table*}

\subsection{MIO formulation description}
We provide Table \ref{tab:formdesc} with short descriptions of the parameters and variables in the MIO formulation of the proposed model from Figure \ref{eq:completeFormulation}. 
\begin{table*}
    \centering
    \caption{Description of MIO symbols used in the proposed formulation in Figure \ref{eq:completeFormulation}. Parameter $n$ refers to the number of samples, $K$ is the number of classes, $p$ is the number of features, and $d$ is the depth of the tree.}
    \label{tab:formdesc}
    \begin{tabular}{lrll}
    \toprule
        \multicolumn{2}{r}{Symbol}  & Explanation & Size \\
    \midrule
    \midrule
        Params & $Y_{ik}$ & Equal 1 for true class of a sample & $n \times K$ \\
        & $\boldsymbol{x}_{i}$ & Input samples & $n \times p$  \\
        & $\boldsymbol{\epsilon}$ & Minimal change in feature values & $p$  \\
        & $\epsilon_{\max}$ & Maximal value of $\boldsymbol{\epsilon}$ & $1$  \\
        & $N_{\min}$ & Minimum of samples in a leaf & $1$  \\
        & $\mathcal{T}_{L}$ & Set of leaf nodes & $2^d$  \\
        & $\mathcal{T}_{B}$ & Set of decision (branching) nodes & $2^d - 1$  \\
        & $A_L(t)$ & Ancestors of leaf $t$ that decide left & $\le d - 1$  \\
        & $A_R(t)$ & Ancestors of leaf $t$ that decide right & $\le d - 1$  \\
    \midrule
        Variables & $Q$ & Tree's leaf accuracy & 1 \\
        & $s_{it}$ & Accuracy potential of $\boldsymbol{x}_i$ in leaf $t$ & $n \times 2^d$ \\
        & $S_{it}$ & Accuracy contribution of $\boldsymbol{x}_i$ in leaf $t$ & $n \times 2^d$ \\
        & $r_{t}$ & Reference accuracy for $s_{:t}$  & $2^d$ \\
        & $z_{it}$ & Assignment of $\boldsymbol{x}_i$ to leaf $t$  & $n \times 2^d$ \\
        & $l_{t}$ & Non-emptiness of leaf $t$  & $2^d$ \\
        & $c_{kt}$ & Assignment of class $k$ to leaf $t$  & $K \times 2^d$ \\
        & $a_{jt}$ & 1 if deciding on feature $j$ in node $t$ & $p \times (2^d - 1)$ \\
        & $b_{t}$ & Decision threshold in node $t$ & $2^d - 1$ \\
    \bottomrule
    \end{tabular}
\end{table*}

\subsection{MIO Solver}
\label{sec:mip_gaps}

We have utilized the Gurobi optimizer as an MIO solver under an academic license. Although the solver makes steady progress towards global optimality, the road there is lengthy. Figure \ref{fig:compare_GAPS_agg} shows the progress of the MIP Gaps during the 8-hour optimization averaged over all datasets. For a detailed, per-dataset view, see Figure \ref{fig:compare_GAPS_detail}. The solution is still improving, albeit rather slowly, after 8 hours. 
The narrowing of the MIP gap is achieved only by finding better feasible solutions.  
This lack of improvement of the objective bound might have been affected by our hyperparameter settings which focused on finding feasible solutions and heuristic search. However, tests with default parameters did not improve the best bound either.

\subsubsection{Default hyperparameters of Gurobi solver}
\label{sec:mip_default}

The performance of the Gurobi optimizer depends on the choice of hyperparameters. 
For the sake of simplicity, we have considered only two sets of parameters.
To measure the performance change of our choice of (hyper)parameters, we ran a test with the default value of the MIPFocus parameter and a test with the default value of the Heuristics parameter. 

\begin{figure*}
  \centering
  \begin{subfigure}[b]{0.9\textwidth}
      \includegraphics[width=\textwidth]{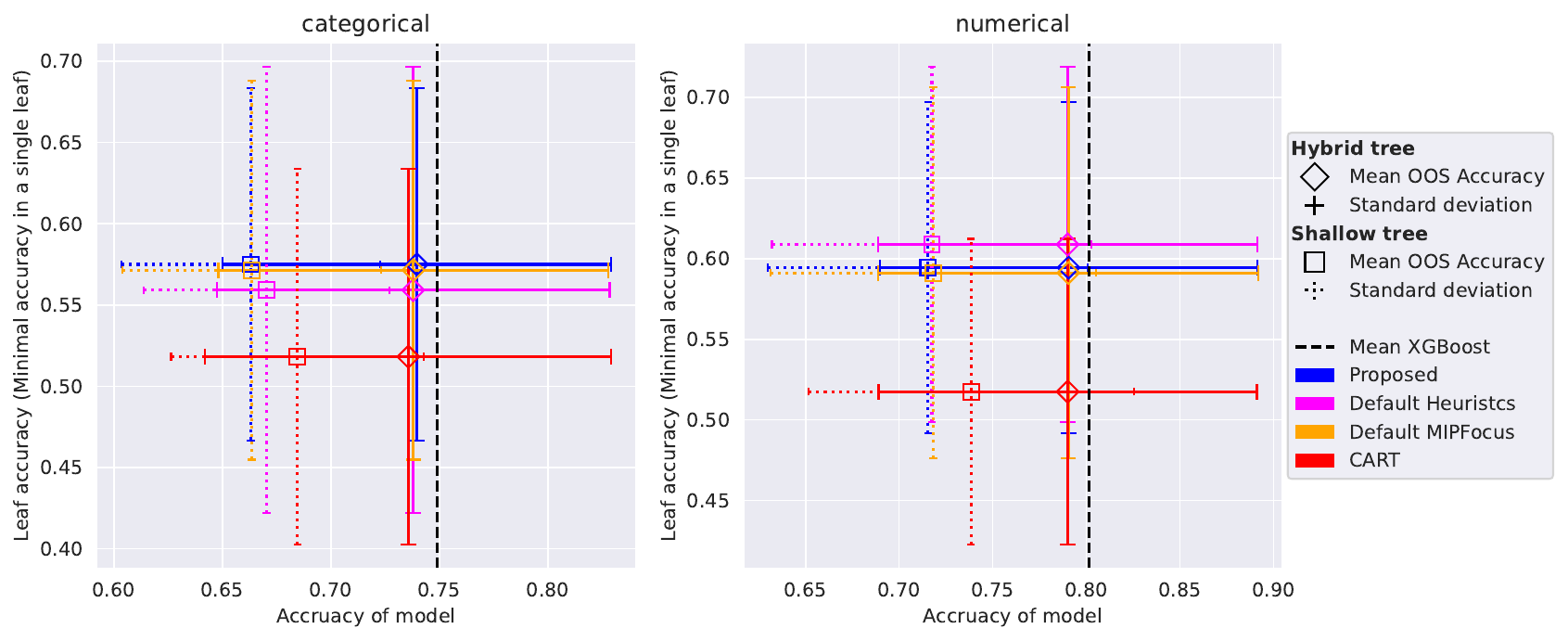}
      \caption{Comparison of the Proposed model to models with default parameter configurations shows varying results. MIPFocus seems to influence the search only very slightly. Heuristics, on the other hand, show significant improvement on numerical datasets and a decrease in performance on categorical datasets, with about the same absolute difference. }
        \label{fig:compare_MIP}    
  \end{subfigure}
  \begin{subfigure}[b]{0.8\textwidth}
  \includegraphics[width=\textwidth]{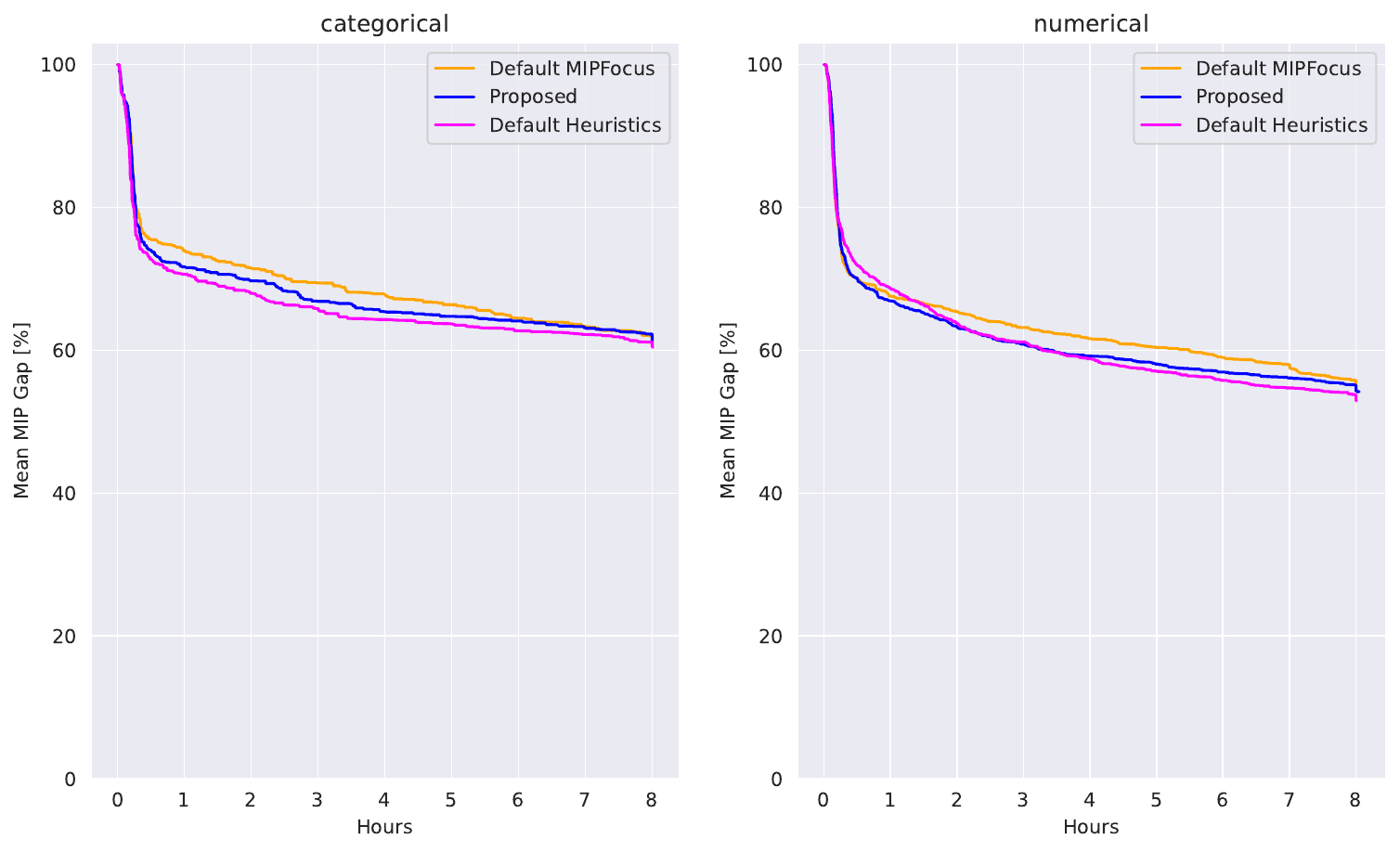}
  \caption{Mean MIO optimality gap development over the solving time, averaged over all datasets. For a non-aggregated version, see Figure \ref{fig:compare_GAPS_detail}.}
    \label{fig:compare_GAPS_agg}
  \end{subfigure}
  \caption{Comparison of models with the proposed configuration of Gurobi hyperparameters and runs with default values of the modified parameters.}
\end{figure*}

\begin{figure*}
    \centering
    \includegraphics[width=\textwidth]{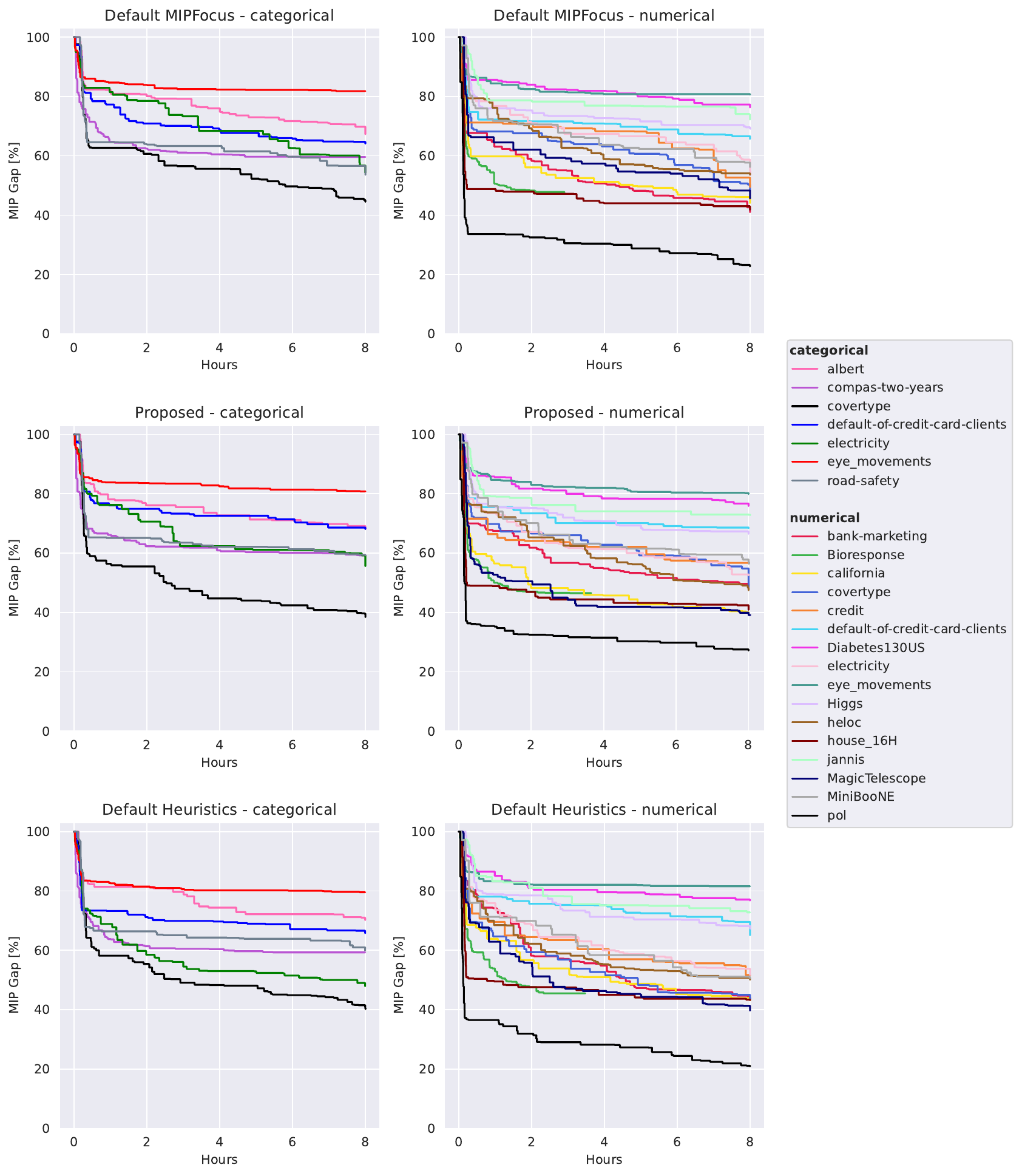}
    \caption{Mean MIO optimality gap development over the solving time averaged over 10 different train-test splits. The figure shows the progress of the value of the MIO optimality gap averaged over all splits of each dataset. Each line corresponds to one dataset. For an aggregated version, see Figure \ref{fig:compare_GAPS_agg}.}
    \label{fig:compare_GAPS_detail}
\end{figure*}

The results (cf. Figure \ref{fig:compare_MIP} and Tables \ref{tab:compare_MIP_h}, \ref{tab:compare_MIP_f}) show no significant improvements regarding the MIOFocus parameter. However, with the default value of the Heuristics parameter, we observe an improvement in performance on numerical datasets and a decrease in performance on categorical datasets. Both absolute differences in accuracy are about 0.015, so we opted for the variant with similar performances on both categorical and numerical datasets. That is the proposed variant focusing on heuristics. This proposed configuration also shows a more stable increase in accuracy w.r.t. the performance of CART models. The solver performance varies per dataset, as visualized in Figure \ref{fig:compare_GAPS_detail}.

These differences in performance suggest that hyperparameter space regarding the MIO solver should be further explored and could yield improvements. A closer look at Figure \ref{fig:compare_GAPS_detail} suggests that different configurations help achieve better conditions for the solver on different datasets. This might be an area of further hyperparameter tuning based on the specific attributes of the dataset.

\begin{table*}
    \centering
    \caption{Detailed view of the differences in the accuracy between the default Heuristics parameter and the proposed configuration (Heuristics = 0.8). A positive number means the accuracy advantage of the proposed hyperparameter configuration. We see absolute mean differences of comparable values. The negative difference in leaf accuracy on numerical datasets also has a higher standard deviation, suggesting a stronger influence by an outlier dataset. For a graphical representation of this data, see Figure \ref{fig:compare_MIP}.}
    \label{tab:compare_MIP_h}
    \begin{tabular}{lrlll}
            \toprule
         & \textbf{Data type} & \textbf{Minimal} & \textbf{Mean ($\pm$ std)} & \textbf{Maximal} \\ 
         \midrule
         \midrule
         \textbf{Leaf Accuracy} & categorical & $-0.0117$ & $0.0158 \pm0.0234$ & $0.0531$ \\
         & numerical & $-0.1178$ & $-0.0143 \pm0.0402$ & $0.0435$ \\
         \midrule
         \textbf{Hybrid-tree Accuracy} & categorical & $-0.0011$ & $0.0017 \pm0.0035$ & $0.0094$ \\
         & numerical & $-0.0047$ & $0.0005 \pm0.0025$ & $0.0062$ \\
         \bottomrule
    \end{tabular}
\end{table*}

\begin{table*}
    \centering
    \caption{Detailed view of the differences in the accuracy between the default MIPFocus parameter and the proposed configuration (MIPFocus = 1). A positive number means the accuracy advantage of the proposed hyperparameter configuration. Both variants seem to perform comparably, with a potential slight edge in favor of the proposed configuration. For a graphical representation of this data, see Figure \ref{fig:compare_MIP}.}
    \label{tab:compare_MIP_f}
    \begin{tabular}{lrlll}
            \toprule
         & \textbf{Data type} & \textbf{Minimal} & \textbf{Mean ($\pm$ std)} & \textbf{Maximal} \\ 
         \midrule
         \midrule
         \textbf{Leaf Accuracy} & categorical & $-0.0304$ & $0.0036 \pm0.0213$ & $0.0299$ \\
         & numerical & $-0.0528$ & $0.0034 \pm0.0342$ & $0.0788$ \\
         \midrule
         \textbf{Hybrid-tree Accuracy} & categorical & $-0.0028$ & $0.0016 \pm0.0043$ & $0.0088$ \\
         & numerical & $-0.0026$ & $0.0001 \pm0.0019$ & $0.0032$ \\
         \bottomrule
    \end{tabular}
\end{table*}

\subsection{Memory requirements}
\begin{figure*}
    \centering
    \begin{subfigure}{0.8\textwidth}
      \includegraphics[width=\textwidth]{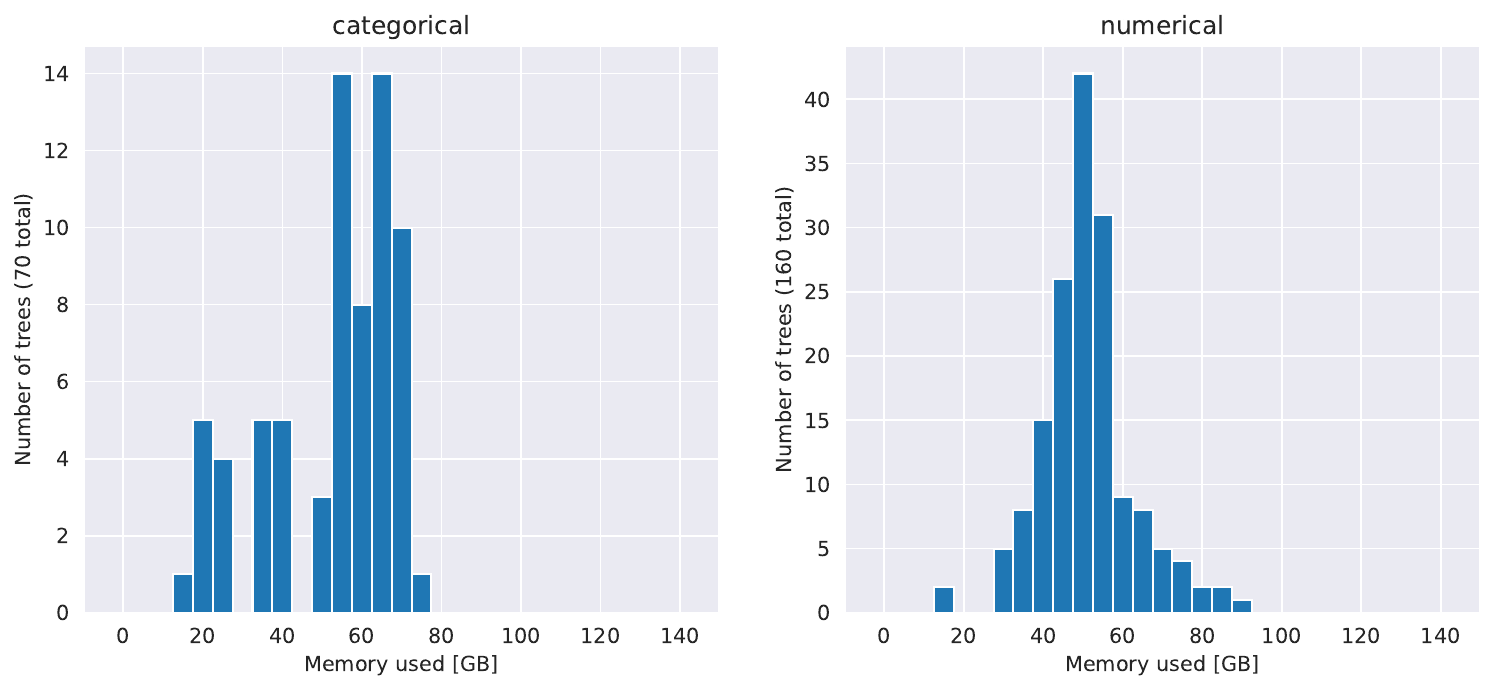}
        \caption{Histogram of memory requirements of MIO solver for all dataset splits.}
    \end{subfigure}
    \begin{subfigure}{0.8\textwidth}
      \includegraphics[width=\textwidth]{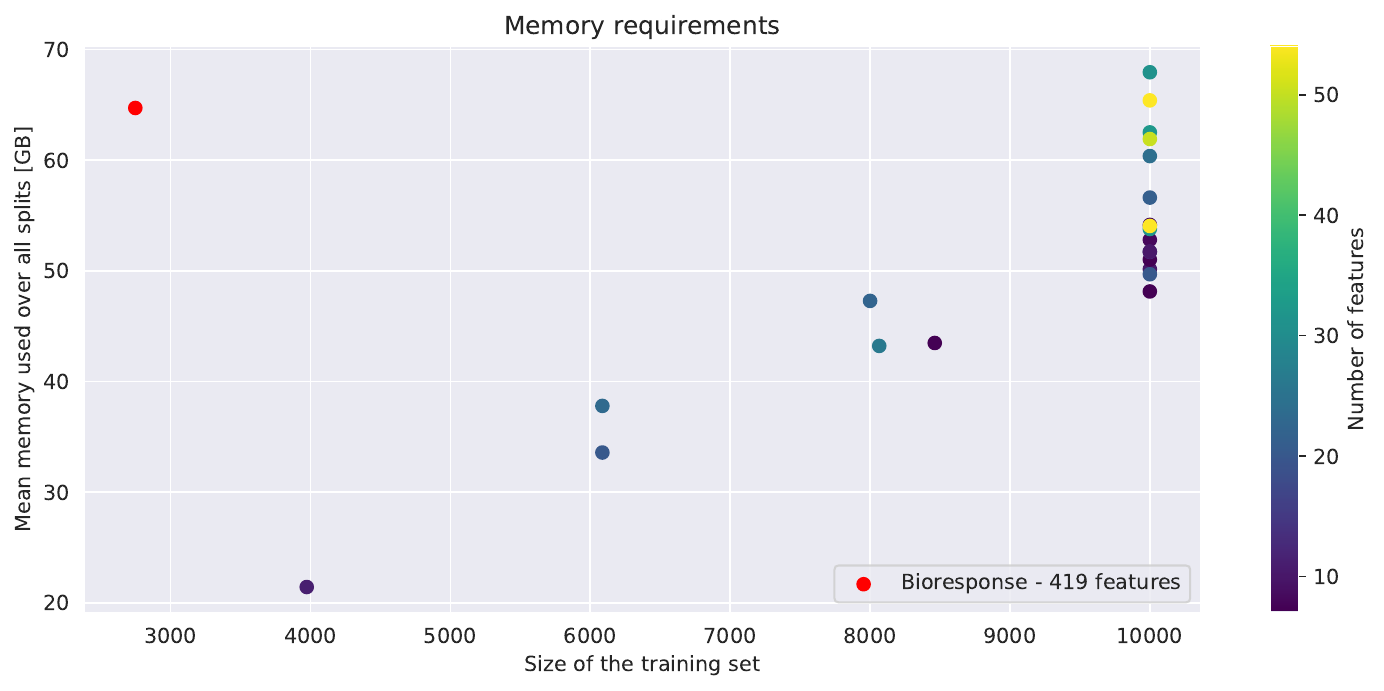}
        \caption{Mean memory requirements on datasets. Dots are colored according to the number of features. Dataset Bioresponse is excluded from the color mapping due to having significantly more features. Training sets were truncated to a maximum of 10,000 points.}
        \label{fig:mem_dependence}
    \end{subfigure}
    \caption{Memory requirements mostly do not exceed 70 GB. The memory requirements increase slightly when more time is given to the solver and significantly increase when bigger training sets are considered. We can also see some correlation between the number of features and memory requirements when looking at same-size datasets.}
    \label{fig:memory}
\end{figure*}

Overall, the memory requirements of the datasets were between 15 and 95 GB. On average, all datasets required at most 70 GB of working memory. Figure \ref{fig:memory} shows the memory requirements of our formulation in more detail. The extension phase of the process is negligible in this regard, as it requires only about 1.5 GB of working memory in total and is performed after the MIO optimization. Training and extending the CART models also required less than 2 GB of working memory.

The amount of memory required by the MIO solver is dependent on the size of the data in the number of training samples, as well as the number of features. Figure \ref{fig:mem_dependence} shows this linear dependence of memory requirements on the size of the training set. Based on the coloring of the nodes, we also see the dependence on the number of features, especially in the case of the Bioresponse dataset.

\subsubsection{Performance of the model given a shorter time}
\begin{figure*}
    \centering
    \begin{subfigure}{.7\textwidth}
      \includegraphics[width=\textwidth]{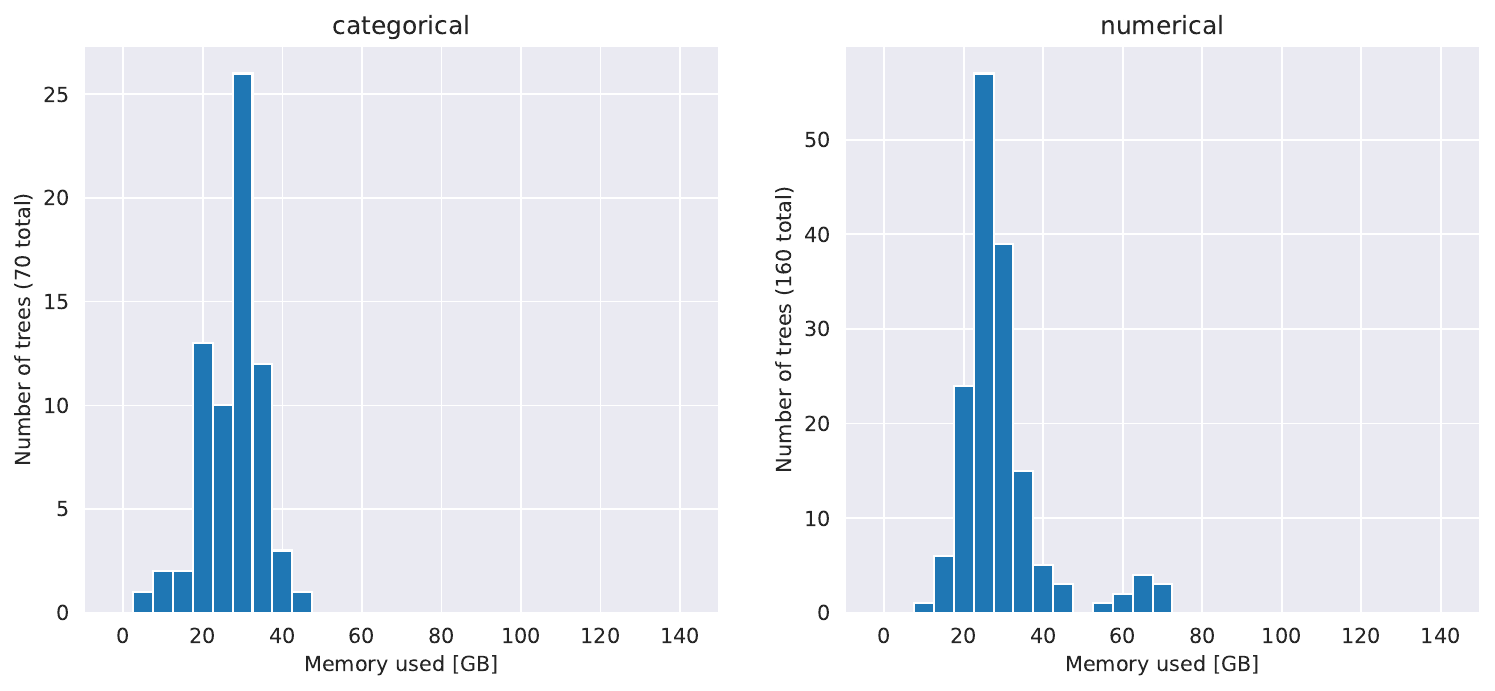}
        \caption{Histogram of memory requirements of MIO solver for all dataset splits.}
    \end{subfigure}
    \begin{subfigure}{0.7\textwidth}
      \includegraphics[width=\textwidth]{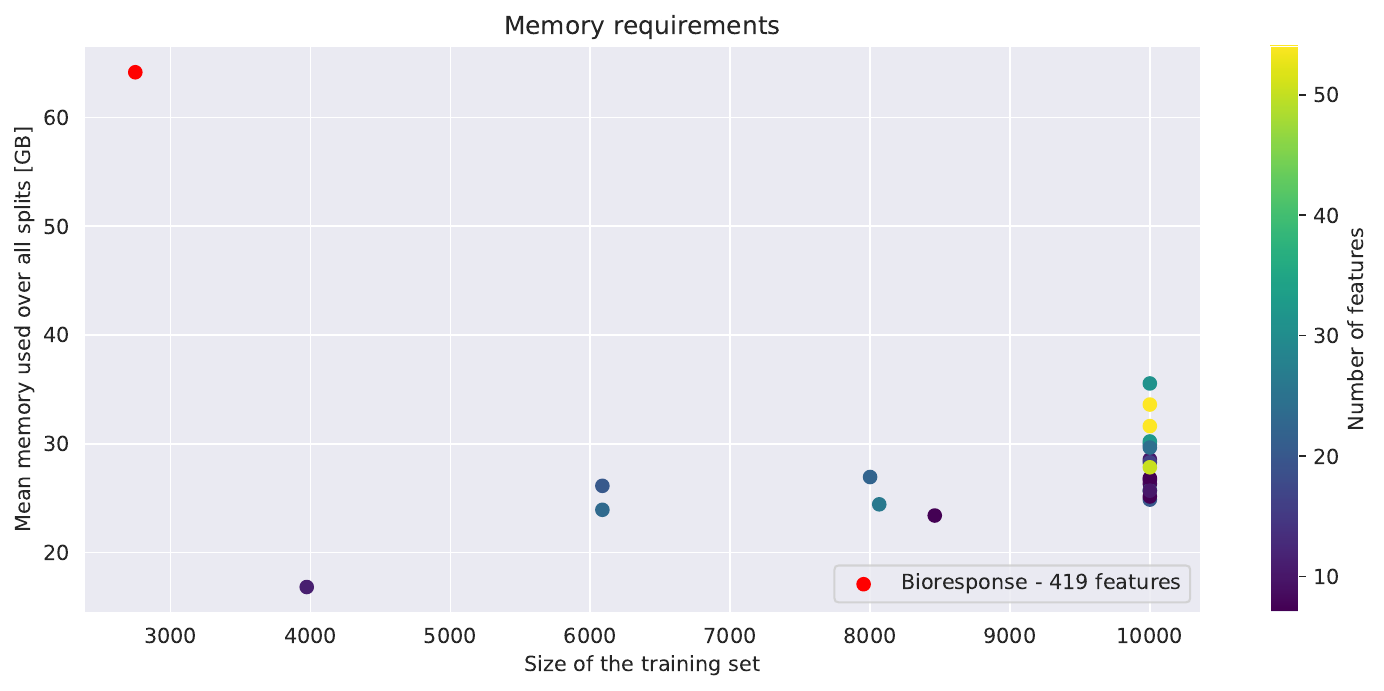}
        \caption{Mean memory requirements on datasets. Dots are colored according to the  number of features. Dataset Bioresponse is excluded from the color mapping due to having a significantly higher number of features. Training sets were clipped to a maximum of 10,000 points.}
    \end{subfigure}
    \begin{subfigure}{0.9\textwidth}
      \includegraphics[width=\textwidth]{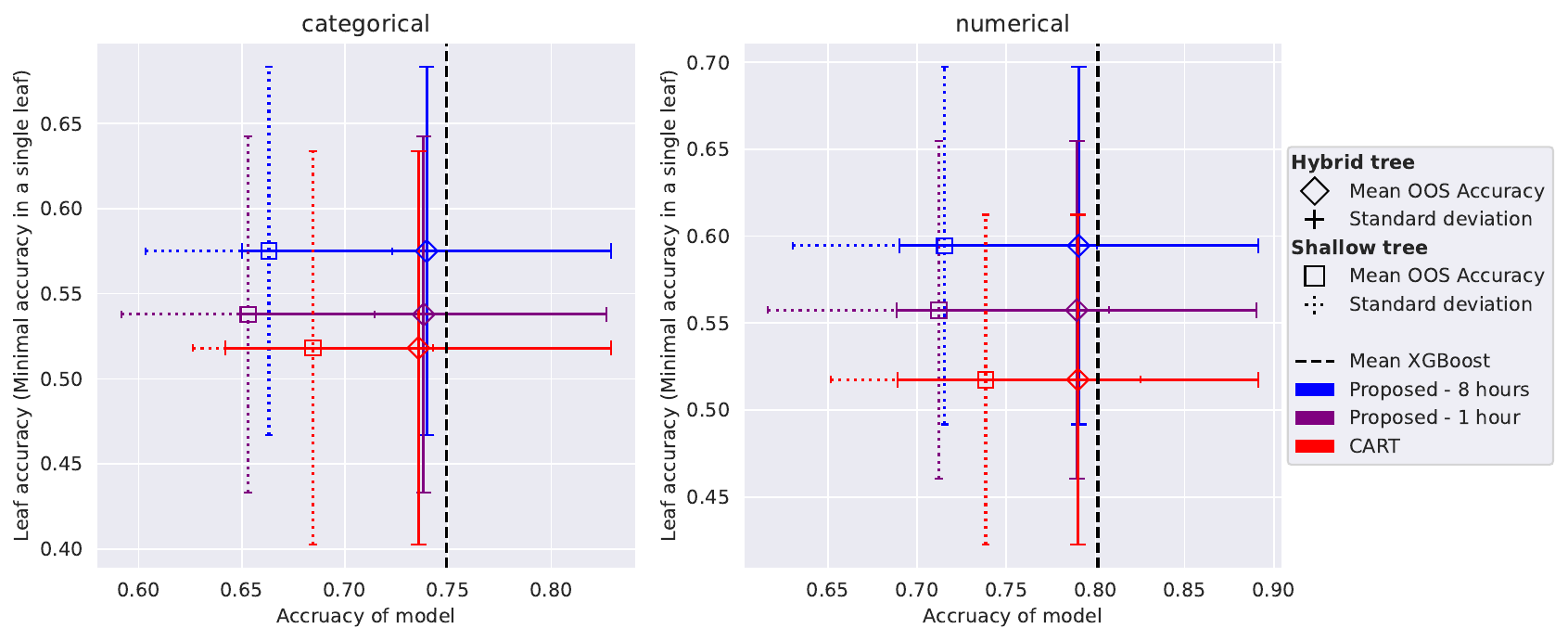}
      \caption{Comparison of the performance of the Proposed model after 1 and 8 hours of optimization.}
      \label{fig:performance_1hour}
    \end{subfigure}
    \caption{Comparison to a version of the Proposed model that the Gurobi solver optimized for only one hour. Compared to the main configuration, which ran for 8 hours, we notice a significant decrease in memory requirements for most datasets, up to tens of gigabytes. An outlier dataset Bioresponse with cca 10 times more features sees a smaller decrease of about 2 GB.}
    \label{fig:memory_1hour}
\end{figure*}

When considering a shorter time for optimization, we can lower the memory requirements to levels attainable by current personal computers. When optimizing our MIO model for one hour, the required memory is below 50 GB for all datasets except Bioresponse, which has one order of magnitude more features than the rest of the datasets included in the benchmark. The mean memory requirement is below 30 GB of working memory (compared to 50 GB for the 8-hour run). See Figure \ref{fig:memory_1hour} for details.

Figure \ref{fig:performance_1hour} shows that even with this limited budget, we can achieve significant improvement compared to CART in leaf accuracy and similar accuracy of hybrid trees.

\subsection{Reduction of the trees}
\begin{figure*}
    \centering
    \includegraphics[width=0.8\textwidth]{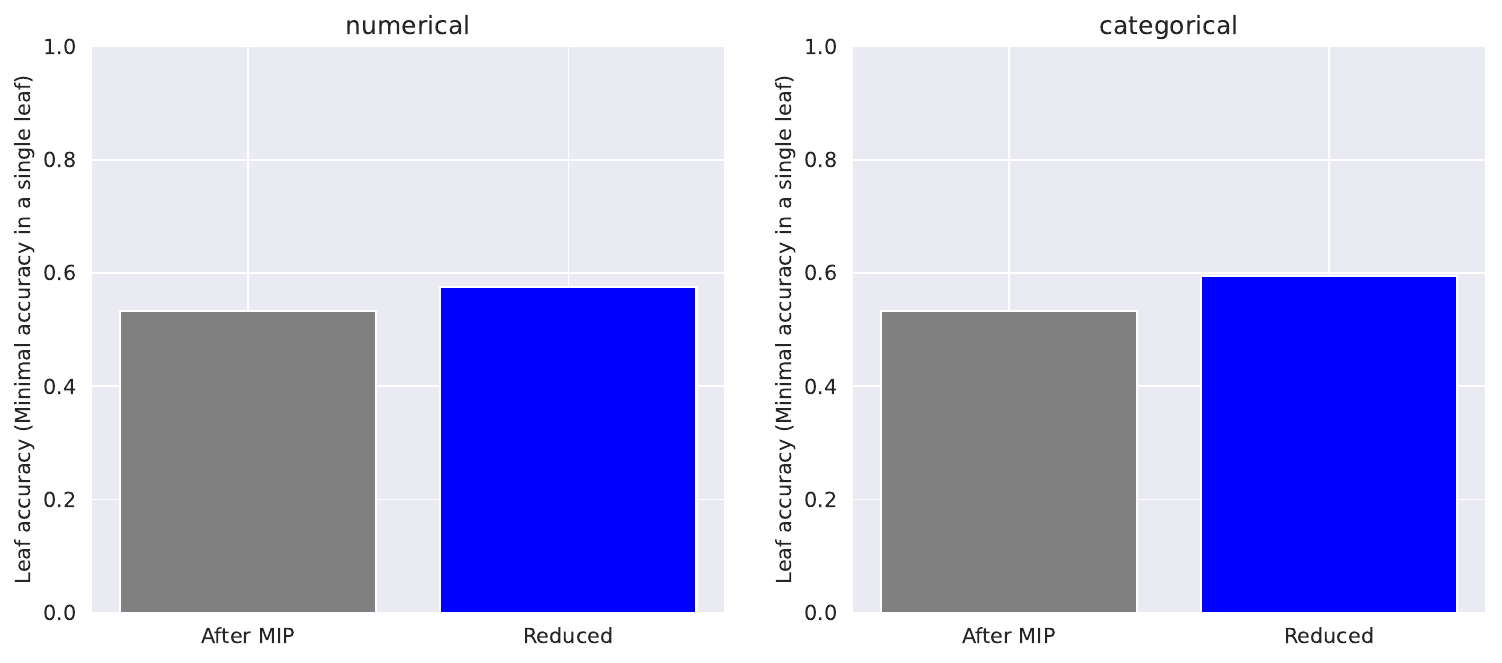}
    \caption{Effect of reduction on leaf accuracy of the model. In grey is the leaf accuracy before reduction, and in blue is the leaf accuracy after reduction. The plot shows mean accuracy over all datasets of a given type created by the proposed model.}
    \label{fig:red_effect}
\end{figure*}

The reduction phase has a beneficial influence on the leaf accuracy of a model. Figure \ref{fig:red_effect} shows this improvement in mean leaf accuracy over all datasets.

\begin{figure*}
  \centering
  \begin{subfigure}[b]{0.75\textwidth}
      \includegraphics[width=\textwidth]{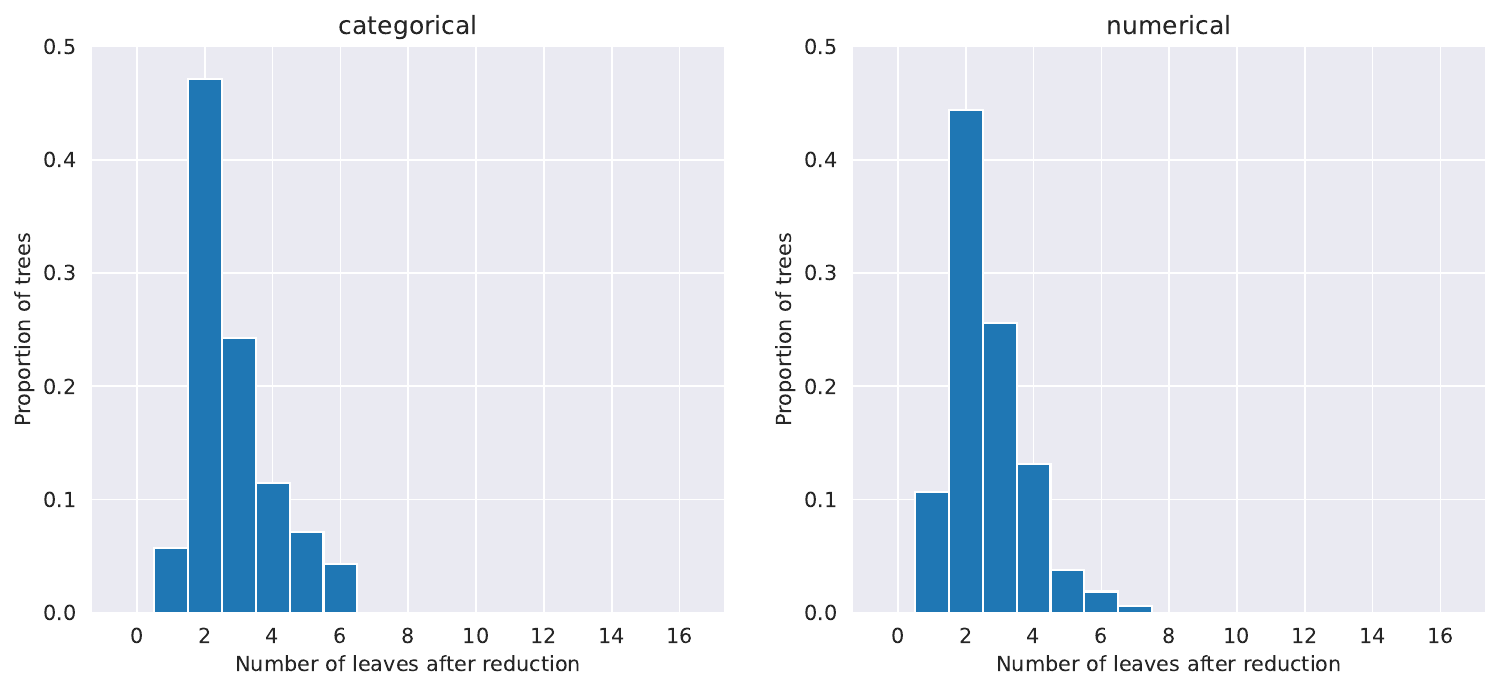}
    \caption{Histogram of the number of leaves of the reduced trees optimized directly using the proposed formulation. The trees are heavily pruned.}
  \end{subfigure}
  \begin{subfigure}[b]{0.75\textwidth}
      \includegraphics[width=\textwidth]{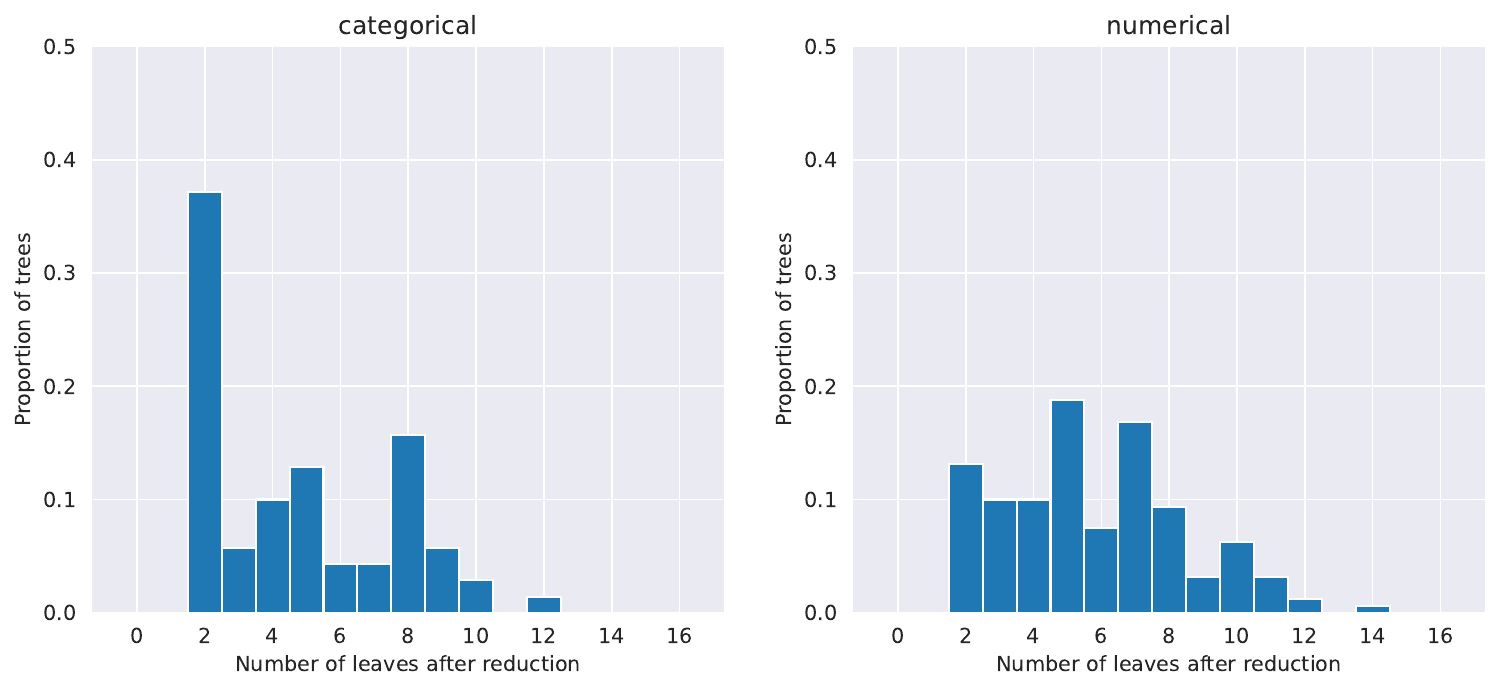}
    \caption{Histogram of the number of leaves of the reduced trees created with the proposed formulation, warmstarted using a simple CART solution. The trees are smaller compared to well-optimized CART but retain some complexity. This was the chosen method.}
  \end{subfigure}
  \begin{subfigure}[b]{0.75\textwidth}
      \includegraphics[width=\textwidth]{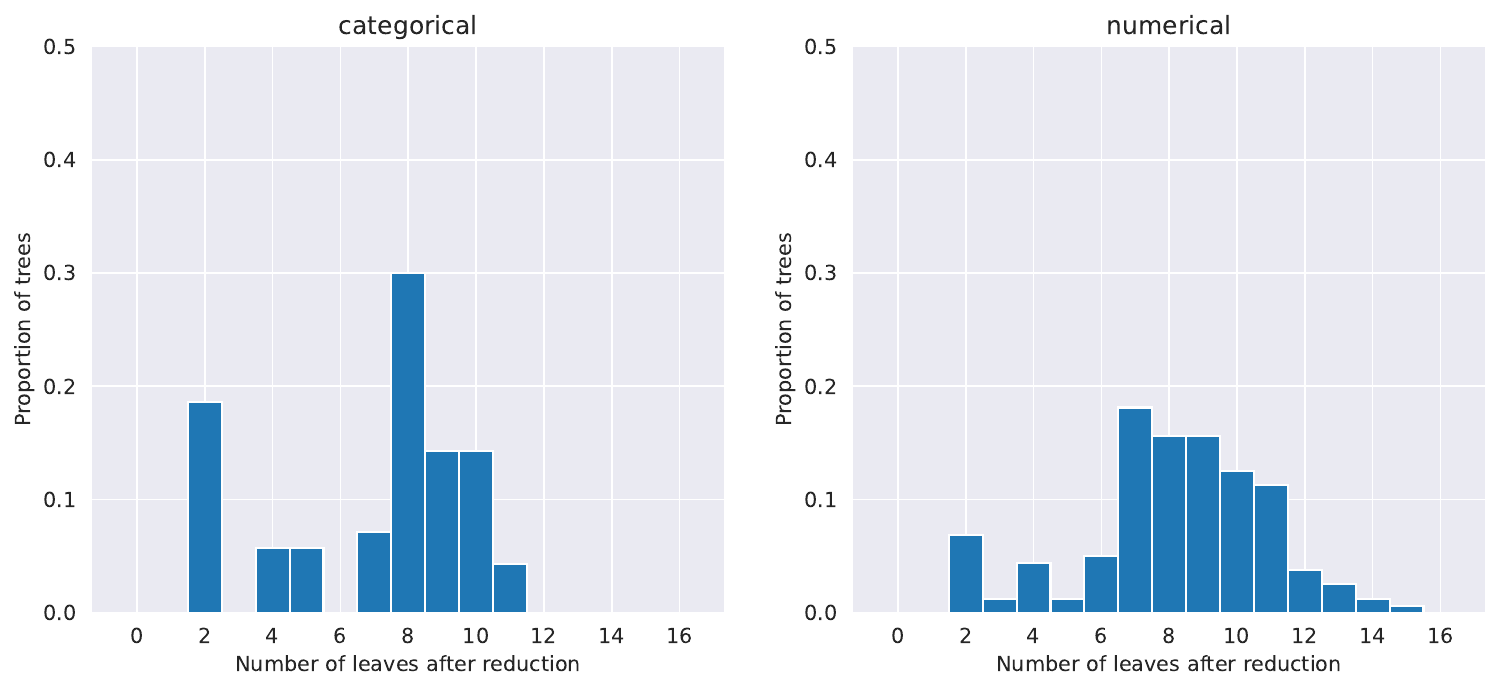}
    \caption{Histogram of the number of leaves of the reduced trees created by CART with optimized hyperparameters.}
  \end{subfigure}
  \caption{Comparison of the numbers of leaves of trees after the reduction procedure.}
    \label{fig:reductions}
\end{figure*}

In Figure \ref{fig:reductions}, we further provide a comparison of the complexity of the created trees by comparing the distributions of the number of leaves (or potential explanations) provided by the method. 

The maximum amount of leaves of a tree with depth 4 is 16. CART model has, on average, around 8 leaves after reduction. The proposed model's distribution is close to the distribution of CART models. When solving the MIO formulation directly, the distribution is severely shifted toward very small trees. Our proposed method uses a default CART solution to warmstart the search, which might explain the shape of the distribution compared to the direct method and CART. 

\subsection{Hyperparameter search distributions}
We needed to optimize hyperparameters for extending models and CART trees used for comparisons. We used Bayesian hyperparameter search for that purpose. 

\subsubsection{Extending XGBoost models}
For the hyperparameter search of XGBoost models in leaves, we used the distributions listed in Table \ref{tab:xgb_distributions}. The parameters are almost all the same as those used by \citep{grinsztajnWhyTreebasedModels2022}. Only the Number of estimators and Max depth were more constrained to account for the fewer samples available for training. 

The Bayesian optimization was run for 50 iterations, with 3-fold cross-validation in every leaf that contained enough points to perform the optimization. The same process was used to extend all tested trees.

\begin{table*}
    \centering
    \caption{Distributions of hyperparameters of extending XGBoost models in leaves. These were used in the Bayesian hyperparameter search in each leaf separately. All distributions except Max depth and Number of estimators are the same as in \citep{grinsztajnWhyTreebasedModels2022}. The two different distributions were selected smaller to improve the optimization time and to account for lower amounts of data.}
    \label{tab:xgb_distributions}
    \begin{tabular}{rl}
         \toprule
         \textbf{Parameter name} & \textbf{Distribution} [range (inclusive)] \\
         \midrule
         \midrule
         Max depth & UniformInteger [1, 7] \\
         Number of estimators & UniformInteger [10, 500] \\
         Min child weight & LogUniformInteger [1, 1e2] \\
         Learning rate & Uniform [1e-5, 0.7] \\
         Subsample & Uniform [0.5, 1] \\
         Col sample by level & Uniform [0.5, 1] \\
         Col sample by tree & Uniform [0.5, 1] \\
         Gamma & LogUniform [1e-8, 7] \\
         Alpha & LogUniform [1e-8, 1e2] \\
         Lambda & LogUniform [1, 4] \\
         \bottomrule
    \end{tabular}
\end{table*}

In leaves with an insufficient amount of samples to perform the cross-validation (less than 3 samples of at least one class in our case), we train an XGBoost model with a single tree of max depth 5. In leaves with 100\% training accuracy, we do not learn any model and use the majority class. 

\subsubsection{CART models}
For the hyperparameter optimization of CART models, we also used Bayesian search, with the distributions shown in Table \ref{tab:cart_distributions}.

\begin{table*}
    \centering
    \caption{Distributions of hyperparameters of CART models used to compare to our method. Max depth and Min samples in a leaf were fixed, but remain in the table for completeness.}
    \label{tab:cart_distributions}
    \begin{tabular}{rl}
         \toprule
         \textbf{Parameter name} & \textbf{Distribution} [range (inclusive)] \\
         \midrule
         \midrule
         Max depth & UniformInteger [4, 4] \\
         Min samples split & UniformInteger [2, 100] \\
         Min samples leaf & UniformInteger [50, 50] \\
         Max leaf nodes & UniformInteger [2, 16] \\  
         Min impurity decrease & Uniform [0, 0.2] \\ 
         Cost complexity pruning parameter $\alpha$ & Uniform [0, 0.3] \\
         \bottomrule
    \end{tabular}
\end{table*}

The search was run for 100 iterations, with 5-fold cross-validation on the same training data sets as our model. After this search, the best hyperparameters were used to train the model on the full training data. The resulting tree was reduced, and every leaf was extended by an XGBoost model in the same way as our models. 


\subsection{Detailed results}
\label{sec:detailed}
\begin{figure*}
  \centering
  \includegraphics[width=\textwidth]{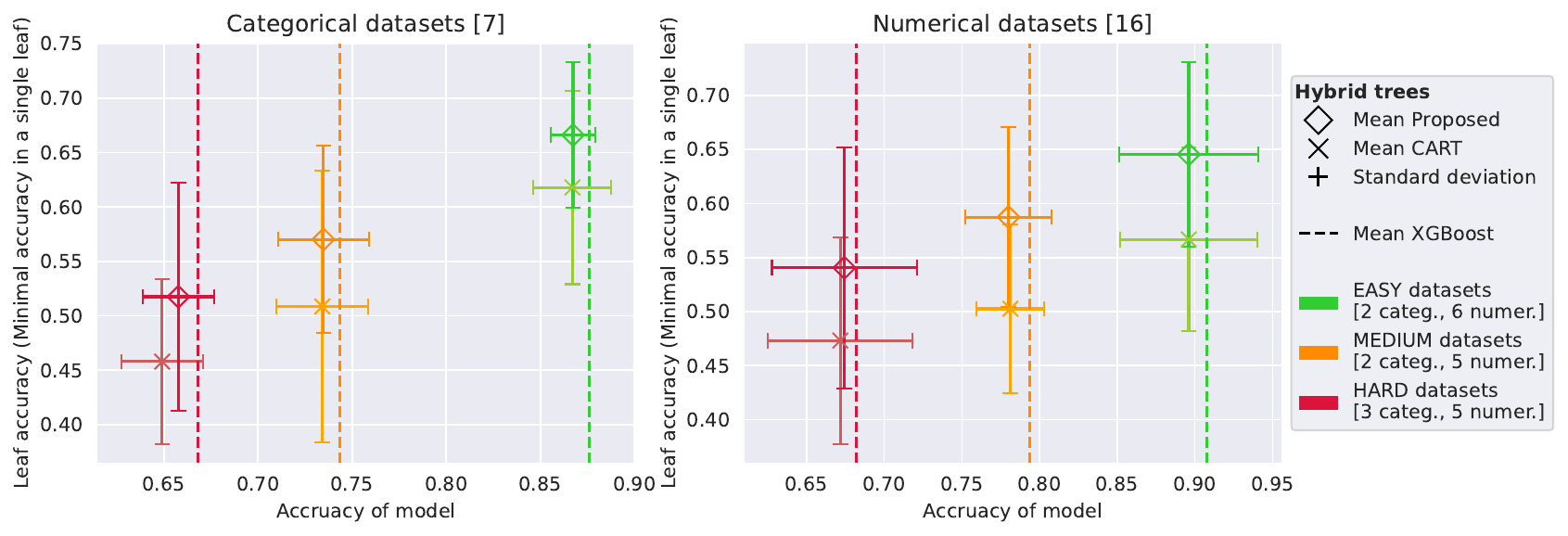}
  \caption{
  Results on out-of-sample data on all classification datasets from the tabular benchmark partitioned into 3 categories by complexity. In square brackets are the numbers of datasets belonging to each partition. This plot shows that our method, when extended in leaves, does not significantly decrease overall performance compared to pure XGBoost while sometimes improving upon accuracy obtainable by extended CART. And it does so consistently for datasets of varying complexity. 
  }
    \label{fig:difficulty}
\end{figure*}

Figure \ref{fig:difficulty} shows again the average performance separately on categorical and numerical datasets divided into three groups by complexity. The complexity measure is based on the performance of XGBoost provided by the benchmark authors. The thresholds of partitions are 0.7 and 0.8 for datasets containing categorical features and 0.75 and 0.85 for datasets with only numerical features. The thresholds were selected in order to separate too easy and too hard datasets, which make the plots less informative, and to explore behavior on datasets with varying inner complexity. We see that the proposed method always significantly improves the leaf accuracy compared to CART.

\begin{figure*}
    \centering
    \includegraphics[width=\textwidth]{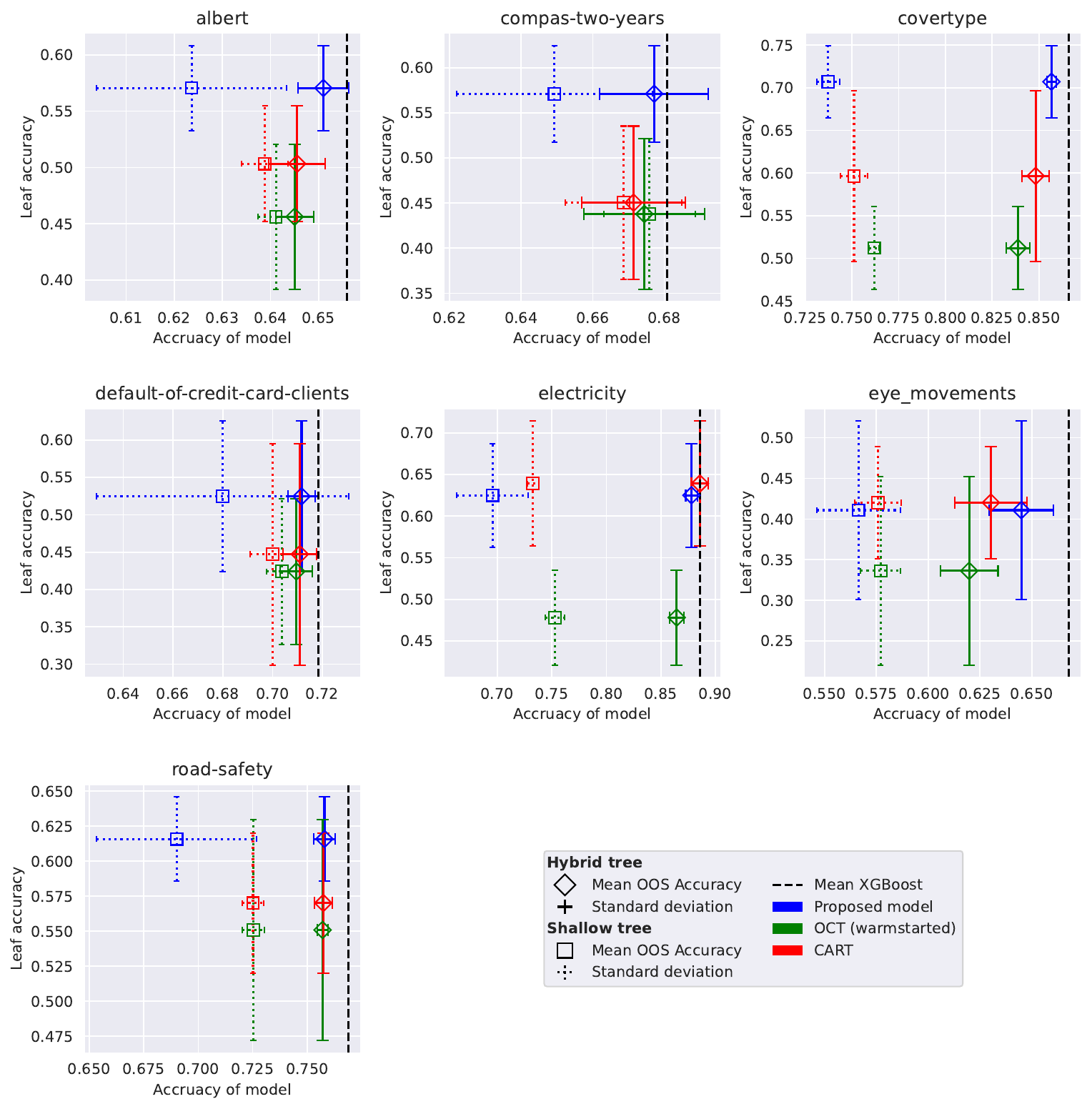}
    \caption{Detailed performance comparison of our model on categorical datasets. 
    }
    \label{fig:categ_detail}
\end{figure*}

\begin{figure*}
    \centering
    \includegraphics[width=\textwidth]{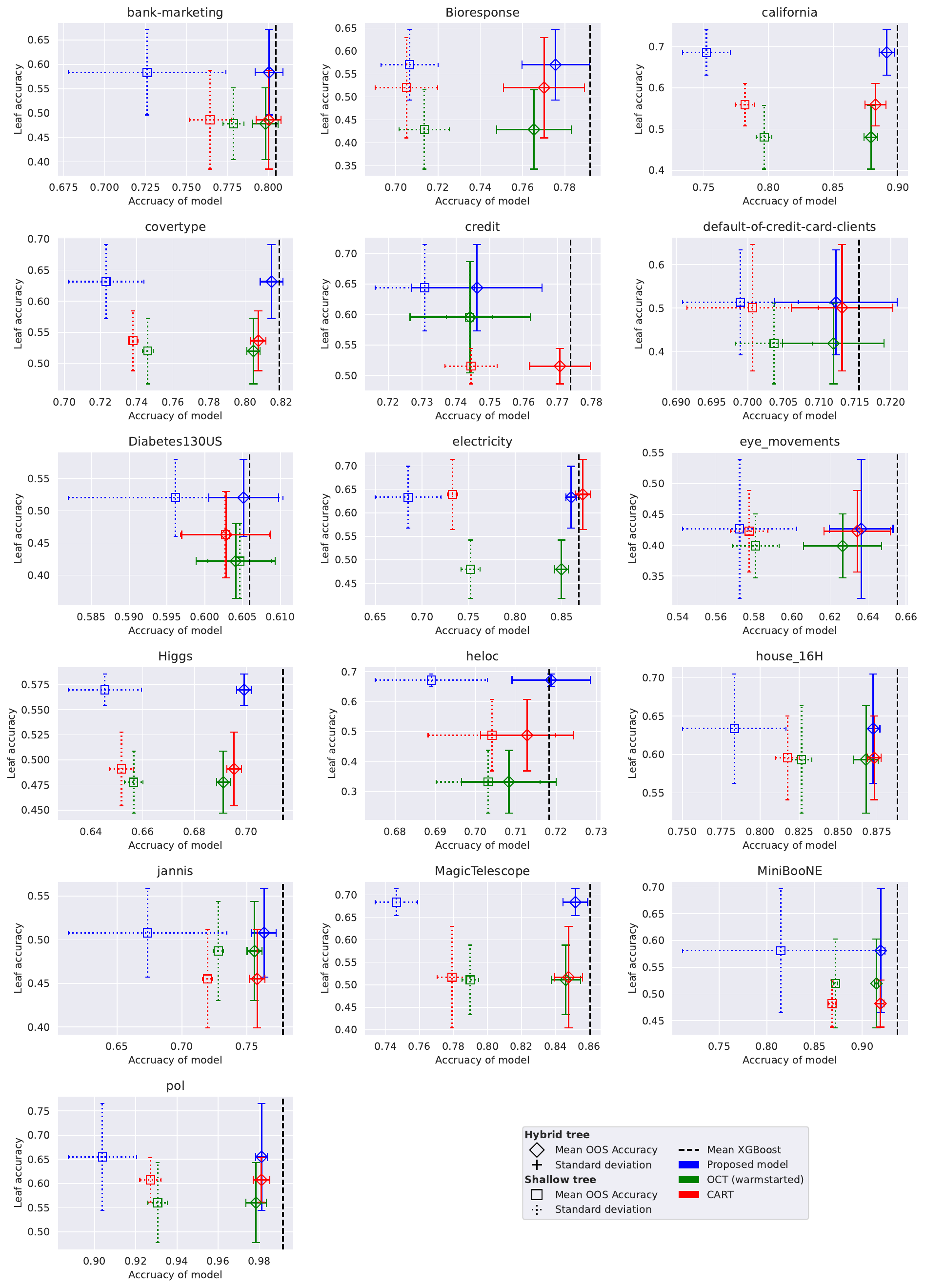}
    \caption{Detailed performance comparison of our model on numerical datasets.}
    \label{fig:numer_detail}
\end{figure*}

We also provide the full results for each dataset. Figures \ref{fig:categ_detail} and \ref{fig:numer_detail} are decomposed variants of Figure \ref{fig:agg} for categorical and numerical datasets, respectively.
We also provide exact results in Tables \ref{tab:categ_detail} and \ref{tab:numer_detail}, respectively. 
The detailed results show that the proposed model outperforms the CART model in both accuracy measures on almost all datasets and has comparable accuracy to XGBoost. Performed statistical tests (signed test and Wilcoxon's signed-rank test) resulted in proving the statistical significance of the better performance of the proposed model, compared to CART.

\begin{table*}
    \centering
    \caption{Categorical datasets. Mean accuracy of models on out-of-sample data and average ranks. }
    \label{tab:categ_detail}
    \begin{tabular}{lrrrrr}
    \toprule
    & \multicolumn{2}{l}{Leaf Accuracy} & \multicolumn{2}{l}{Hybrid-tree Accuracy} & \\
    \cmidrule(lr){2-3} \cmidrule(lr){4-5}
    \textbf{categorical datasets} &   CART &  Proposed &   CART &  Proposed &  XGBoost \\
    \cmidrule(lr){1-3} \cmidrule(lr){4-6}
    albert                         & 0.5033 &    \textbf{0.5706} & 0.6455 &    0.6510 &   \textbf{0.6559} \\
    compas-two-years               & 0.4504 &    \textbf{0.5711} & 0.6714 &    0.6772 &   \textbf{0.6807} \\
    covertype                      & 0.5966 &    \textbf{0.7071} & 0.8482 &    0.8567 &   \textbf{0.8658} \\
    default-of-credit-card-clients & 0.4471 &    \textbf{0.5246} & 0.7110 &    0.7117 &   \textbf{0.7184} \\
    electricity                    & \textbf{0.6392} &    0.6250 & 0.8859 &    0.8781 &   \textbf{0.8861} \\
    eye\_movements                  & \textbf{0.4202} &    0.4109 & 0.6303 &    0.6449 &   \textbf{0.6677} \\
    road-safety                    & 0.5701 &    \textbf{0.6158} & 0.7573 &    0.7579 &   \textbf{0.7689} \\
    \cmidrule(lr){1-3} \cmidrule(lr){4-6}
    Mean rank                      & 1.7143 &    \textbf{1.2857} & 2.8571 &    2.1429 &   \textbf{1.0000} \\
    \bottomrule
    \end{tabular}
\end{table*}

\begin{table*}
    \centering
    \caption{Numerical datasets. Mean accuracy of models on out-of-sample data and average ranks. }
    \label{tab:numer_detail}
    \begin{tabular}{lrrrrr}
    \toprule
    & \multicolumn{2}{l}{Leaf Accuracy} & \multicolumn{2}{l}{Hybrid-tree Accuracy} & \\
    \cmidrule(lr){2-3} \cmidrule(lr){4-5}
    \textbf{numerical datasets} &   CART &  Proposed &   CART &  Proposed &  XGBoost \\
    \cmidrule(lr){1-3} \cmidrule(lr){4-6}
    bank-marketing                 & 0.4861 &    \textbf{0.5837} & 0.8001 &    0.8003 &   \textbf{0.8044} \\
    Bioresponse                    & 0.5201 &    \textbf{0.5700} & 0.7702 &    0.7755 &   \textbf{0.7920} \\
    california                     & 0.5593 &    \textbf{0.6861} & 0.8827 &    0.8914 &   \textbf{0.8997} \\
    covertype                      & 0.5365 &    \textbf{0.6314} & 0.8074 &    0.8147 &   \textbf{0.8190} \\
    credit                         & 0.5153 &    \textbf{0.6439} & 0.7707 &    0.7462 &   \textbf{0.7738} \\
    default-of-credit-card-clients & 0.5011 &    \textbf{0.5136} & 0.7132 &    0.7124 &   \textbf{0.7156} \\
    Diabetes130US                  & 0.4630 &    \textbf{0.5204} & 0.6028 &    0.6051 &   \textbf{0.6059} \\
    electricity                    & \textbf{0.6392} &    0.6331 & \textbf{0.8724} &    0.8600 &   0.8683 \\
    eye\_movements                  & 0.4229 &    \textbf{0.4265} & 0.6343 &    0.6364 &   \textbf{0.6554} \\
    Higgs                          & 0.4910 &    \textbf{0.5698} & 0.6953 &    0.6992 &   \textbf{0.7142} \\
    heloc                          & 0.4881 &    \textbf{0.6722} & 0.7128 &    \textbf{0.7188} &   0.7183 \\
    house\_16H                      & 0.5956 &    \textbf{0.6336} & 0.8733 &    0.8726 &   \textbf{0.8881} \\
    jannis                         & 0.4550 &    \textbf{0.5079} & 0.7579 &    0.7632 &   \textbf{0.7778} \\
    MagicTelescope                 & 0.5168 &    \textbf{0.6835} & 0.8478 &    0.8518 &   \textbf{0.8605} \\
    MiniBooNE                      & 0.4821 &    \textbf{0.5809} & 0.9192 &    0.9194 &   \textbf{0.9369} \\
    pol                            & 0.6073 &    \textbf{0.6550} & 0.9810 &    0.9811 &   \textbf{0.9915} \\
    \cmidrule(lr){1-3} \cmidrule(lr){4-6}
    Mean rank                      & 1.9375 &    \textbf{1.0625} & 2.6875 &    2.1875 &   \textbf{1.1250} \\
    \bottomrule
    \end{tabular}
\end{table*}

\subsection{Other optimization approaches}
\begin{figure*}
  \centering
  \includegraphics[width=\textwidth]{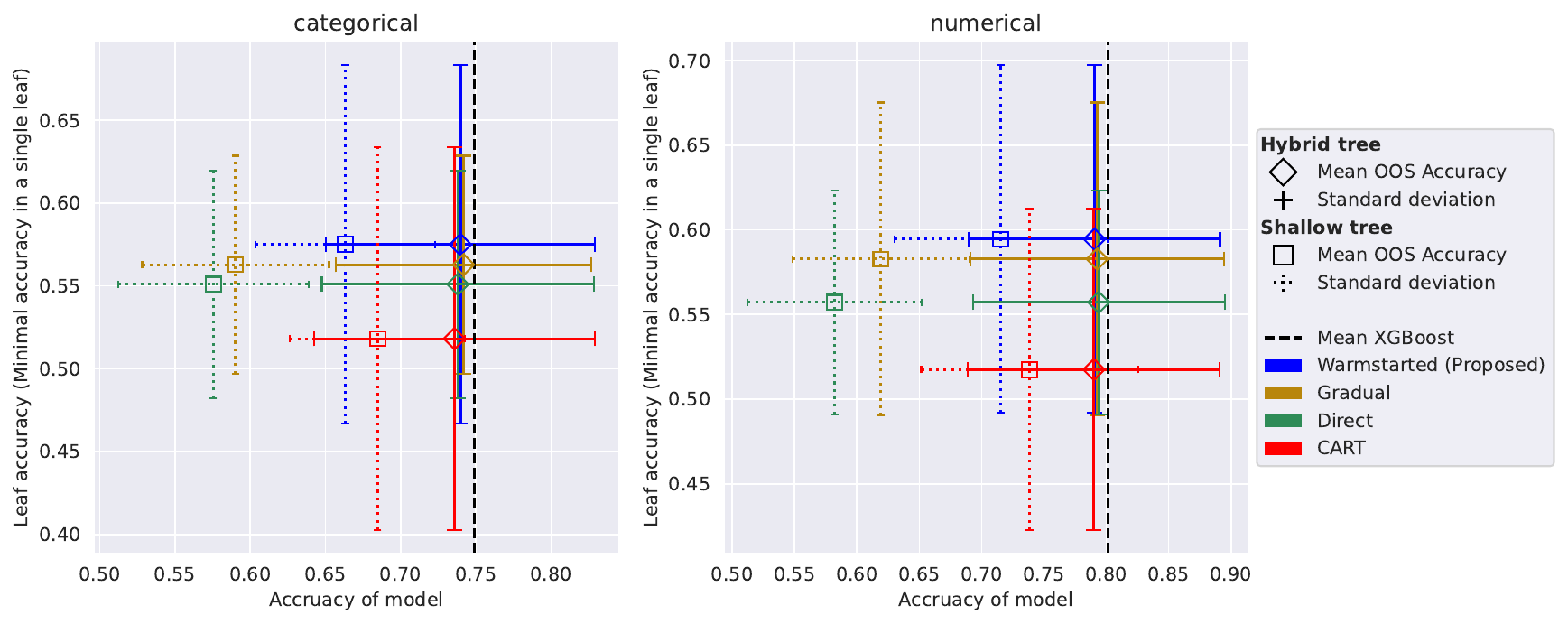}
  \caption{Comparison of the various approaches to the optimization given the same resources and conditions. Warmstarted refers to the approach of starting the optimization process with a CART solution. The gradual approach is the approach of increasing the depth of a tree and starting each new depth with the solution of the previous, shallower tree. Direct means a simple, straightforward optimization of the formulation, as it is stated, without any hints. All three approaches were run with the same resources. For a closer investigation of the Gradual approach, see Table \ref{tab:compare_gradual}.}
    \label{fig:full_comparison}
\end{figure*}

The best-performing approach of warmstarting the MIO solver with a CART solution is not the only one we tested. In Figure \ref{fig:full_comparison}, we see a comparison of three different approaches to optimization. 
\begin{itemize}
    \item \textit{Direct} refers to the straightforward use of the MIO formulation.
    \item \textit{Warmstarted} uses a simple CART solution (created using default hyperparameters) as a starting point of the solving process.
    \item \textit{Gradual} refers to a special process where we start by training a tree with a depth equal to 1 and use the solution found in some given time to start the search for a tree with a depth of 2, and so forth until we reach the desired depth.
\end{itemize}

All three approaches were run with the same resources. This meant that even the gradual approach took 8 hours in total. The time was distributed in a way that the available time for the optimization process doubled with each increase in depth. This means 32 minutes for the first run, 64 minutes for the tree of depth 2, 128 for depth 3, and 4 hours 16 minutes for the final tree with depth 4. 

Interestingly, while the direct approach understandably does not reach a performance similar to the warmstarted variant, the gradual approach shows more promise. It has higher hybrid-tree accuracy by another 0.2 percentage points on average while having lower leaf accuracy by about 1.2 percentage points compared to the warmstarted approach (cf. Table \ref{tab:compare_gradual}).

\begin{table*}
    \centering
    \caption{Comparison of Gradual and Warmstarted approach. Positive numbers show an advantage in the mean accuracy of the Proposed (Warmstarted) approach. Gradual refers to the approach when the depth of the tree is gradually increased during the optimization process.}
    \label{tab:compare_gradual}
    \begin{tabular}{lrlll}
            \toprule
         & \textbf{Data type} & \textbf{Minimal} & \textbf{Mean ($\pm$ std)} & \textbf{Maximal} \\ 
         \midrule
         \midrule
         \textbf{Leaf Accuracy} & categorical & $-0.1094$ & $0.0122 \pm0.0753$ & $0.1130$ \\
         & numerical & $-0.0867$ & $0.0117 \pm0.0624$ & $0.1154$ \\
         \midrule
         \textbf{Hybrid-tree Accuracy} & categorical & $-0.0219$ & $-0.0021 \pm0.0094$ & $0.0083$ \\
         & numerical & $-0.0103$ & $-0.0023 \pm0.0056$ & $0.0076$ \\
         \bottomrule
    \end{tabular}
\end{table*}


\subsection{Ablation Analyses}
We provide some comparing experiments performed by changing a single hyperparameter (or a few related ones) and comparing the performance.

\subsubsection{Unlimited depth CART}
\label{sec:unlim_cart}
\begin{figure*}
  \centering
  \includegraphics[width=\textwidth]{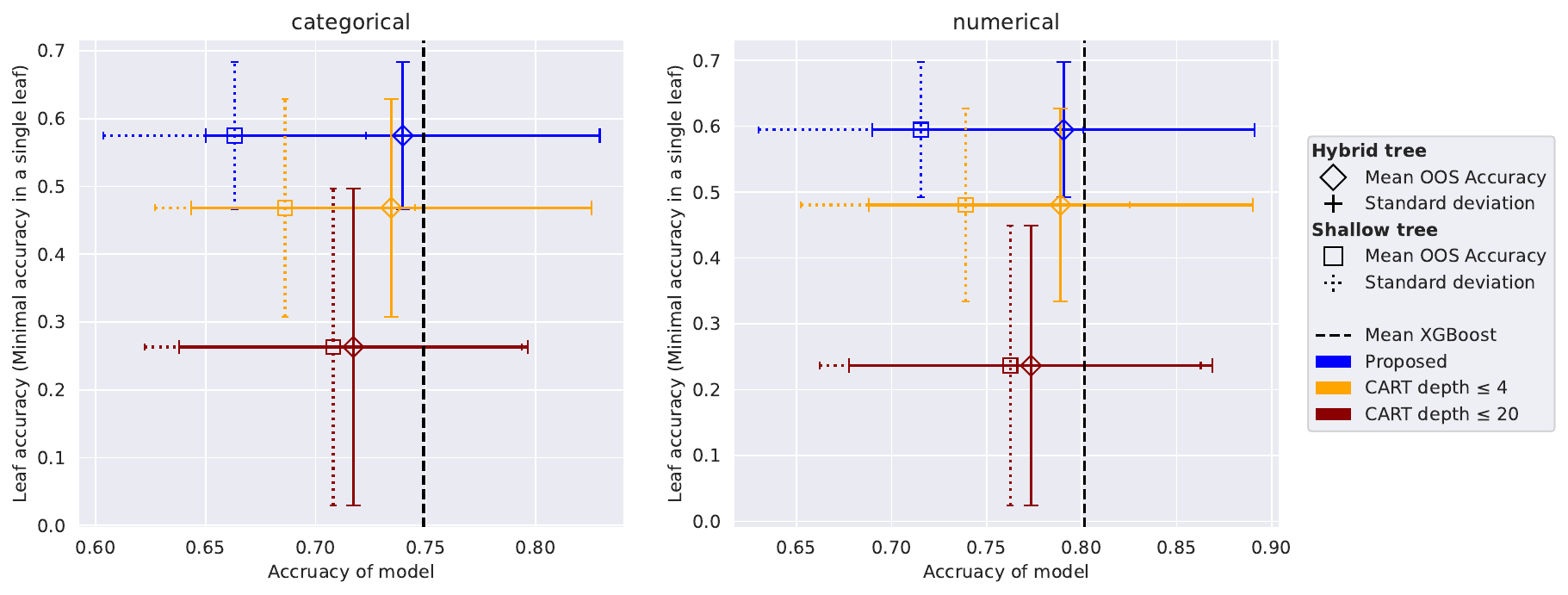}
  \caption{Comparison of CART tree results with limited depth and without such a strict limit on the depth. Deeper trees provide worse explanations (due to the length of explanation) and perform worse in both accuracy measures. For a more detailed description of the differences introduced by the depth, consult Table \ref{tab:cart_depth}.}
    \label{fig:cart_depth}
\end{figure*}

An argument could be made against our choice to compare our method to CART trees with the same limit on depth. Figure \ref{fig:cart_depth} and Table \ref{tab:cart_depth} in more detail show a comparison of CART models with a maximal depth of 4 and a maximal depth of 20. The actual depth limit for each model was optimized along with other hyperparameters using the Bayes hyperparameter optimization procedure. 

Note that these tests were performed in earlier stages of testing without a fixed lower bound on the number of samples in a leaf and without cost complexity pruning. The lower bound on the number of samples was optimized using the Bayes optimization in the range [0, 50]. 

The aggregated results show worse performance regarding both leaf accuracy and hybrid-tree accuracy. Not only do the deeper trees perform worse, but the length of provided explanations is also well above the 5-9 threshold suggested as the limit of human understanding \citep{feldman2000minimization}.

\begin{table*}
    \centering
    \caption{Detailed view of the differences in the accuracy between CART trees with max depth 4 and CART trees with max depth 20. A positive number means the accuracy advantage of the more constrained model (depth $\le 4$). For a graphical representation, see Figure \ref{fig:cart_depth}.}
    \label{tab:cart_depth}
    \begin{tabular}{lrlll}
            \toprule
         & \textbf{Data type} & \textbf{Minimal} & \textbf{Mean ($\pm$ std)} & \textbf{Maximal} \\ 
         \midrule
         \midrule
         \textbf{Leaf Accuracy} & categorical & $-0.0769$ & $0.2053 \pm0.2389$ & $0.5404$ \\
         & numerical & $-0.1183$ & $0.2441 \pm0.2115$ & $0.5680$ \\
         \midrule
         \textbf{Hybrid-tree Accuracy} & categorical & $-0.0025$ & $0.0173 \pm0.0185$ & $0.0420$ \\
         & numerical & $-0.0006$ & $0.0156 \pm0.0119$ & $0.0370$ \\
         \bottomrule
    \end{tabular}
\end{table*}

\subsubsection{Different minimum number of samples in leaves}
\label{sec:Nmin}
\begin{figure*}
  \centering
  \includegraphics[width=\textwidth]{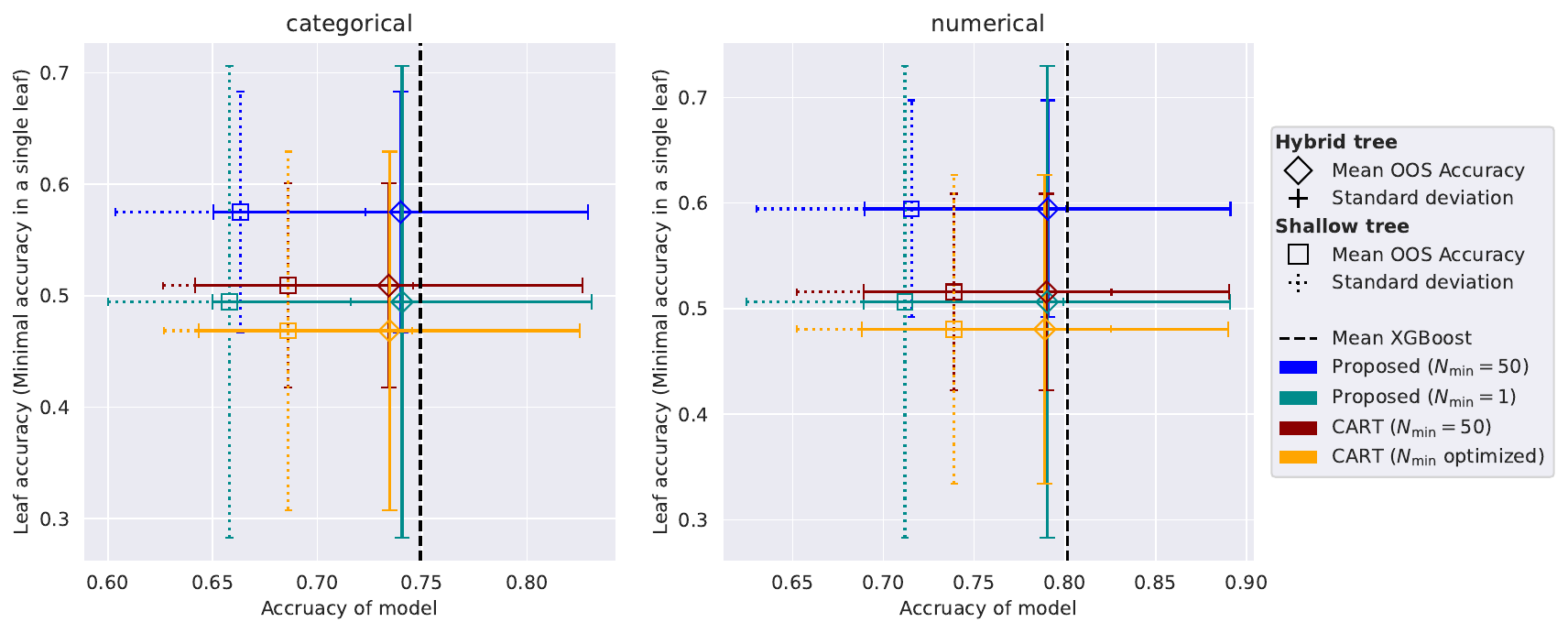}
  \caption{Comparison of performance of the proposed model with minimum samples in leaves equal 50 and 1 and of CART trees with parameter equivalent to $N_{\min}$ fixed to 50 or undergoing hyperparameter optimization. Low values of $N_{\min}$ lead to overfitting to training data and worse out-of-sample performance. Notice the high deviation of the model with $N_{\min} = 1$. CART trees suffer from a similar thing, which suggests two things. Hyperparameter optimization does not opt for high $N_{\min}$ values, and this seems to be a property of trees in general, no matter how they are obtained.}
    \label{fig:compare_nolim}
\end{figure*}

A similar comparison is to see the performance of classically optimized lower bound on the number of samples in each leaf. Figure \ref{fig:compare_nolim} shows a comparison of CART models when the lower bound is fixed to 50 and when it is optimized within the range from 1 to 60 using Bayesian hyperparameter optimization. The figure also includes the performance of the proposed model when $N_{\min}$ is set to 1.

It stresses the importance of setting a minimal amount of samples in leaves. Without enough points to support the leaf's accuracy, it is more likely to be overfitted. On the other hand, when choosing the $N_{\min}$ parameter too high, we restrict some possibly beneficial splits, supported by a smaller amount of training data. 

$N_{\min}$ is a critical hyperparameter, and further testing could provide more insight into the proposed model's performance.

\subsubsection{Non-warmstarted OCT}
\begin{figure*}
  \centering
  \includegraphics[width=\textwidth]{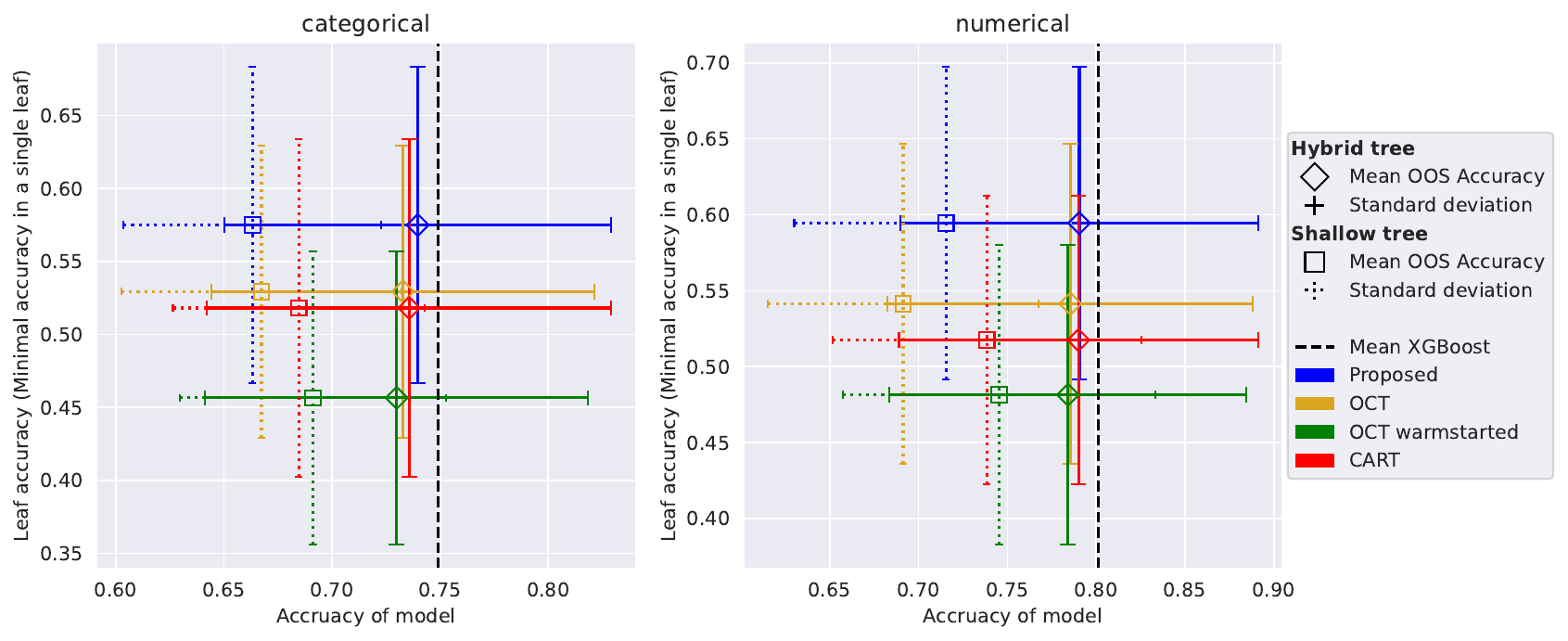}
  \caption{Comparison of OCT trees that are warmstarted the same way as our Proposed model and OCT without the warmstart, optimized directly. Interestingly, direct OCT performs significantly better.}
    \label{fig:compare_OCT}
\end{figure*}

We compare our method to warmstarted OCT because the proposed method also starts from the same initial CART solution. This makes them more comparable. However, we also tested the OCT variant directly optimized from the MIO formulation. See the results in Figure \ref{fig:compare_OCT}. Both OCT models were run with the same hyperparameters as the proposed model. Those being the heuristics-oriented solver, depth equal to 4, and a minimal amount of samples in leaves equal to 50. 

\begin{figure*}
    \centering
    \begin{subfigure}[b]{\textwidth}
        \centering
        \includegraphics[width=0.9\textwidth]{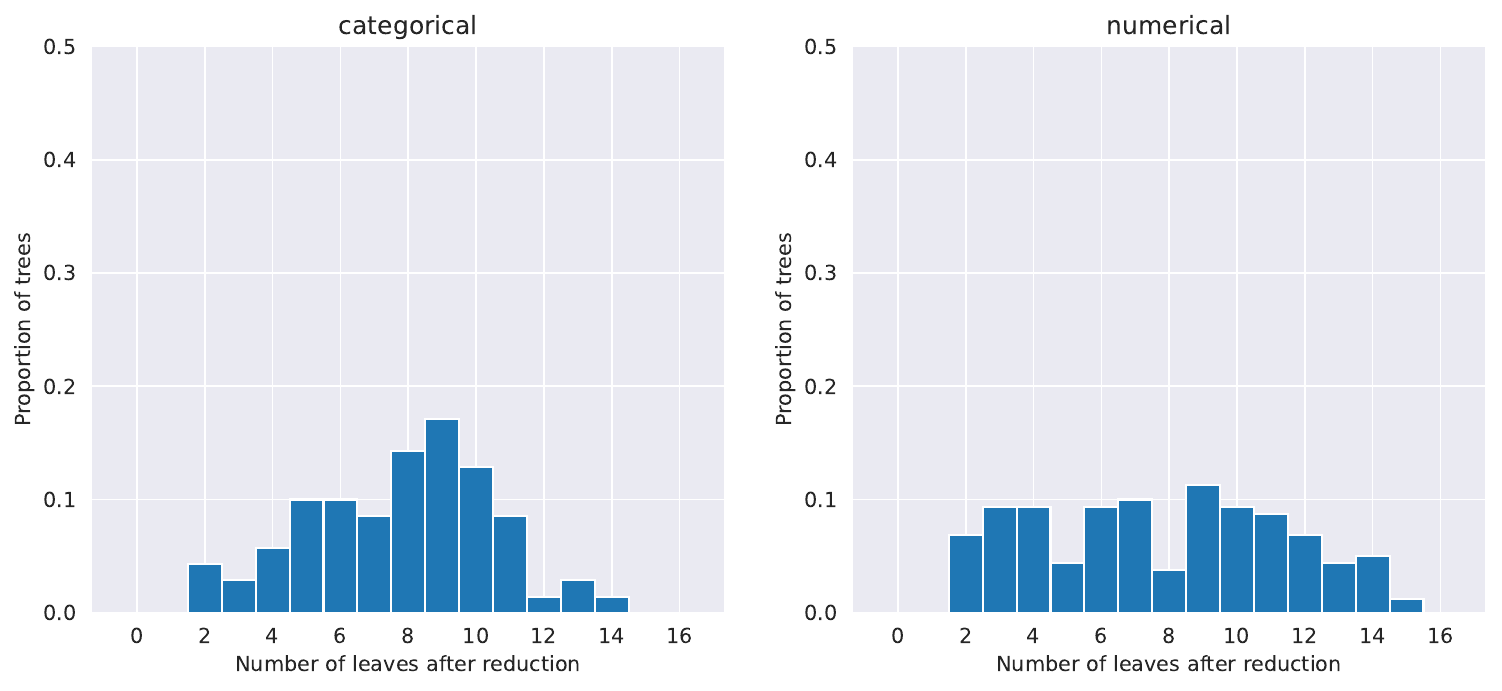}     
        \caption{Histogram of the number of leaves in the reduced trees of the direct OCT method}   
    \end{subfigure}
    \begin{subfigure}[b]{\textwidth}
        \centering
        \includegraphics[width=0.9\textwidth]{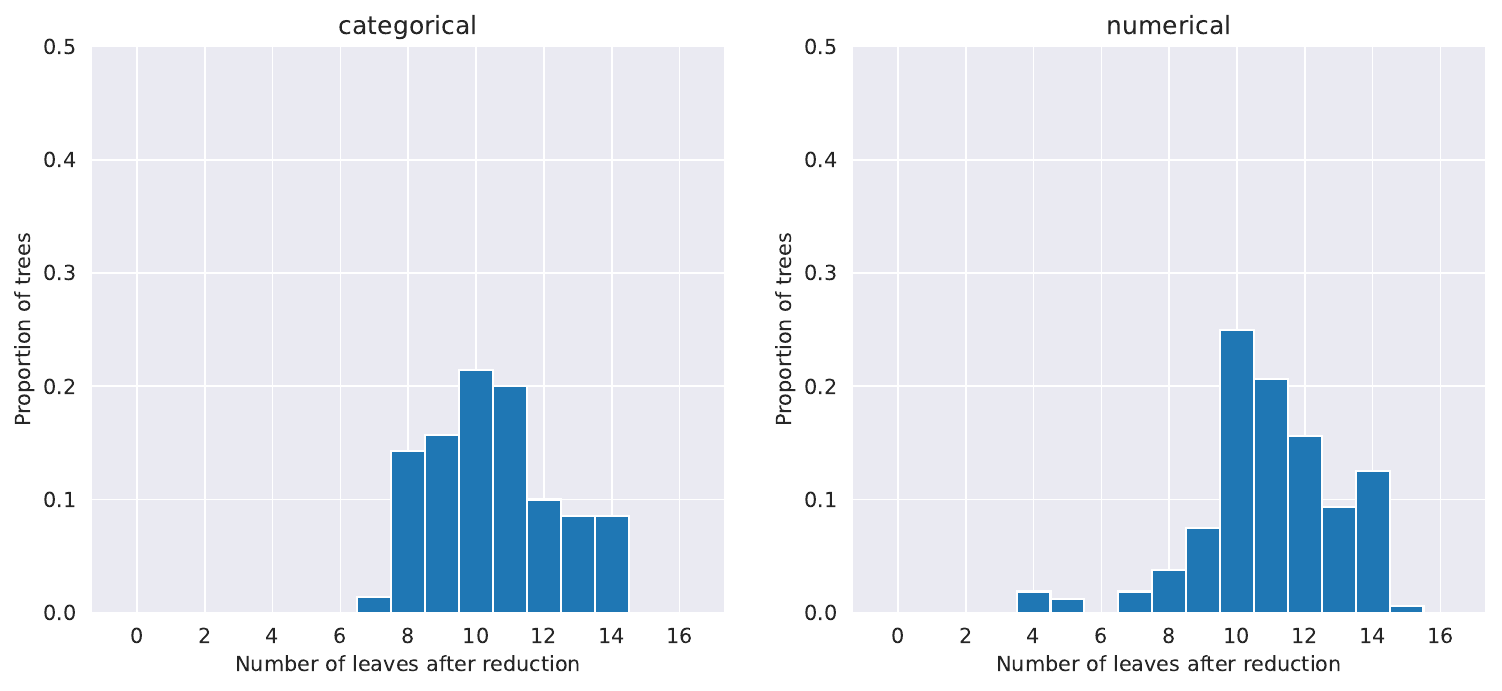}
        \caption{Histogram of the number of leaves in the reduced trees of the warmstarted OCT method.}      
    \end{subfigure}
    \caption{Comparison of reduced tree complexity of the OCT with and without warmstart. OCT without warmstart creates trees of similar distribution as the CART method (cf. Figure \ref{fig:reductions}). And it achieves better leaf accuracy than CART (cf. Figure \ref{fig:compare_OCT}) despite neither optimizing that objective.}
    \label{fig:reductions_OCT}
\end{figure*}

The average OCT performs worse than all our approaches (cf. Figure \ref{fig:full_comparison}, all approaches are above the 0.55 mark, contrary to OCT in Figure \ref{fig:compare_OCT}), but the improvement from the warmstarted variant is intriguing since it clearly manages to overtake the CART model. Especially considering that it is not caused by the direct OCT method's inability to create complex trees without warmstarting. This is supported by Figure \ref{fig:reductions_OCT} showing a distribution of the number of leaves similar to the distribution of CART trees (cf. Figure \ref{fig:reductions}). This suggests that the OCT trees have comparable tree complexity to CART and provide more valid explanations than CART, even without our extension to the formulation. This is an interesting result, considering the fact that neither CART nor OCT methods optimize for leaf accuracy.

Our model, however, more than doubles the improvement of direct OCT. 

\subsubsection{Deeper trees}
\begin{figure*}
    \centering
 
    \begin{subfigure}{0.7\textwidth}
      \centering
      \includegraphics[width=\textwidth]{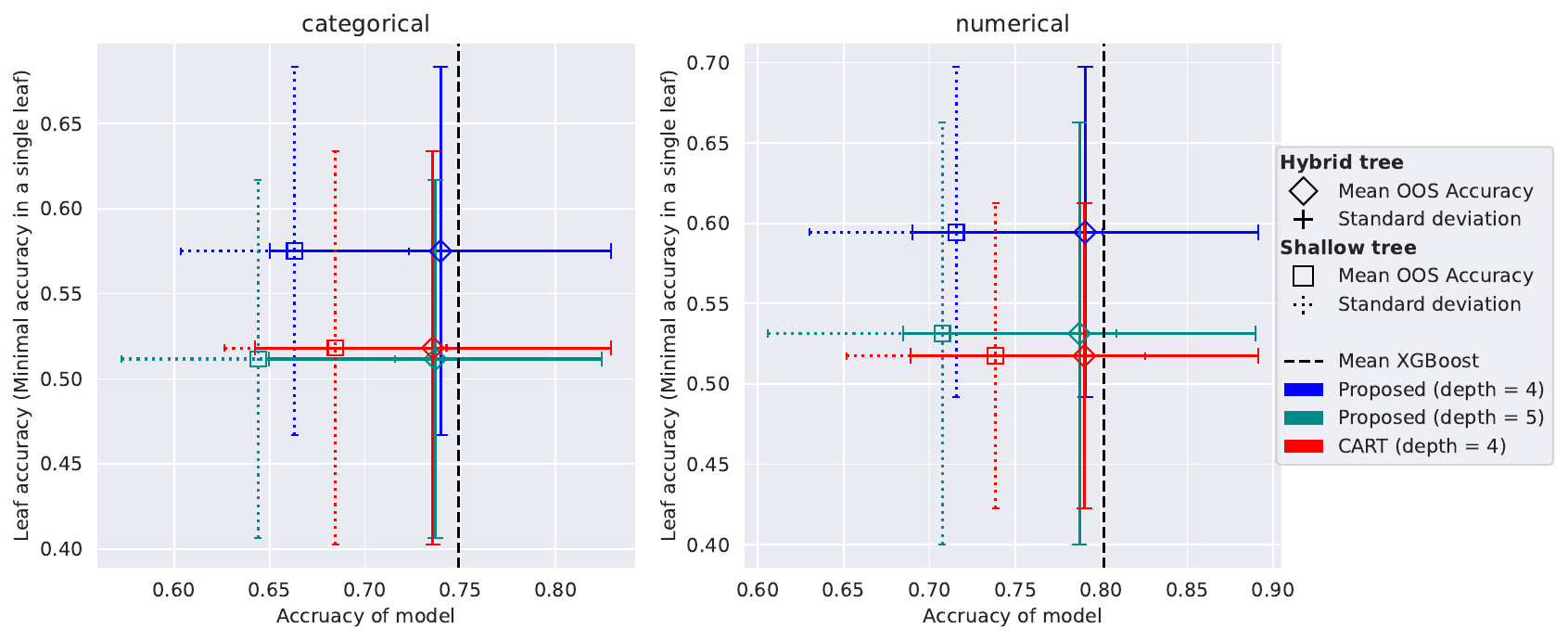}
      \caption{Comparison of performances of the proposed model with depths 4 and 5. Shallower trees perform better, possibly because they are easier to optimize.}
        \label{fig:compare_depth}
    \end{subfigure}
    
    \begin{subfigure}{0.7\textwidth}
      \centering
      \includegraphics[width=\textwidth]{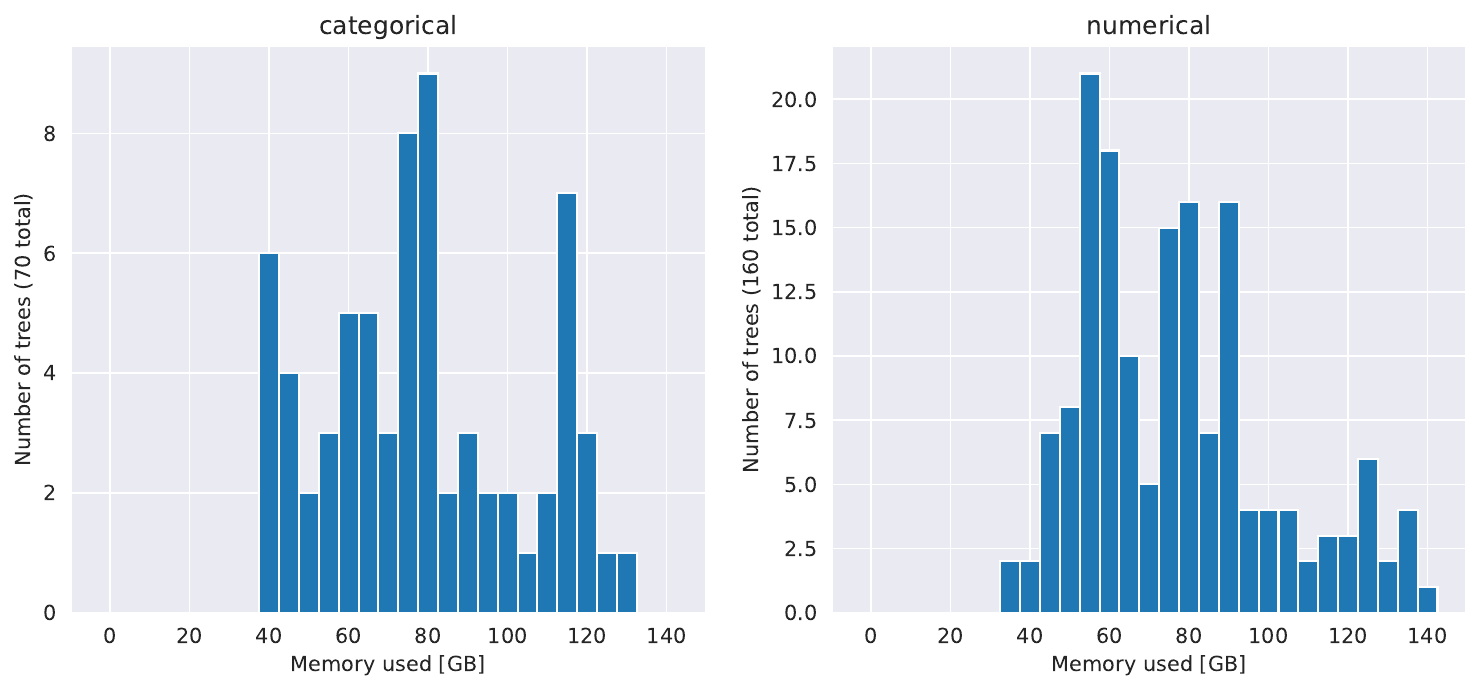}
        \caption{Depth 5. Histogram of memory requirements of MIO solver for all dataset splits.}
    \end{subfigure}
    \begin{subfigure}{\textwidth}
        \centering
      \includegraphics[width=0.7\textwidth]{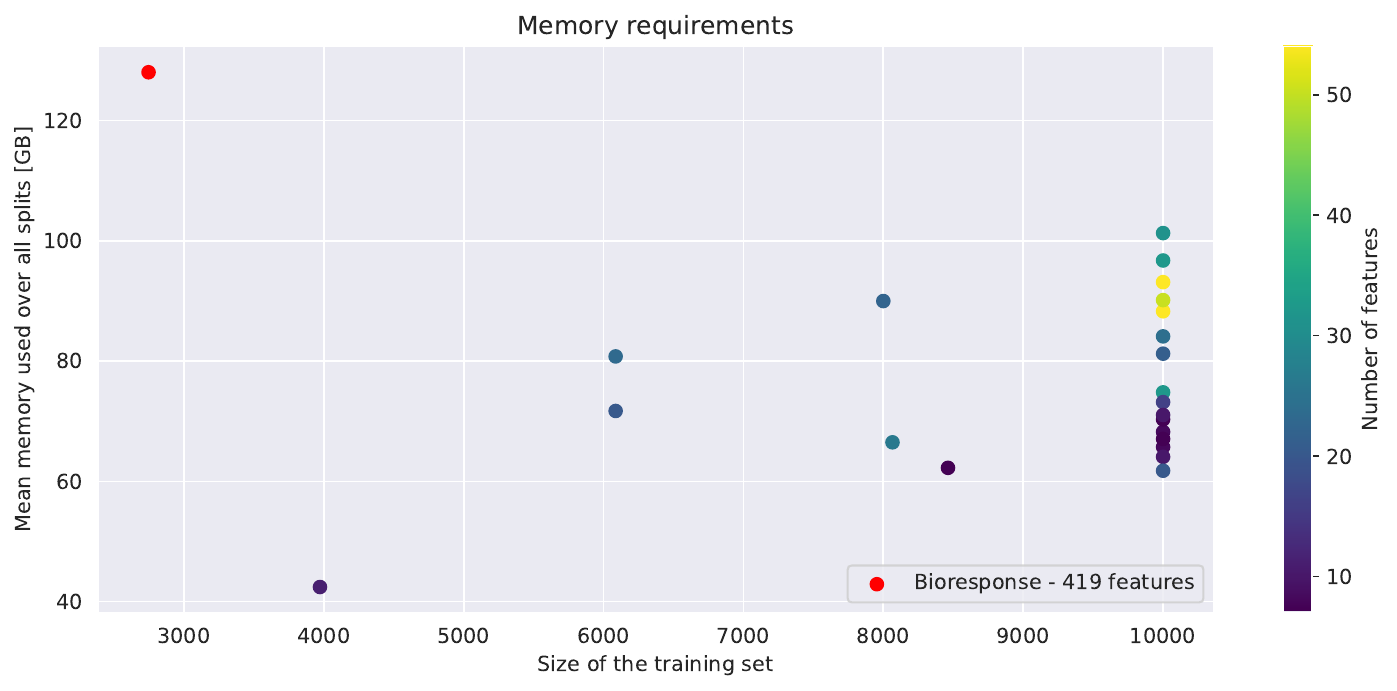}
        \caption{Depth 5. Mean memory requirements on datasets. Dots are colored according to the number of features. Dataset Bioresponse is excluded from the color mapping due to having a significantly higher number of features. Training sets were clipped to a maximum of 10,000 points.}
    \end{subfigure}
    \caption{Comparison of the memory requirements of the Proposed model with depth 5. The mean memory requirement almost increases from cca 51.1 GB to 77.3 GB with an increase in depth from 4 to 5. Compare the above plots with Figure \ref{fig:memory}.}
    \label{fig:memory_depth}
\end{figure*}

Lastly, we provide a comparison of the proposed model of depths 4 and 5. Figure \ref{fig:compare_depth} shows better overall results for shallower trees. This is likely caused by the exponential increase in memory requirements, given the decrease in overall accuracy as well. We provide data about its memory usage in Figure \ref{fig:memory_depth}.
With a model of twice the complexity, the solver struggles to achieve comparable results to the shallower proposed model. 

This is certainly a topic of further exploration by incorporating scalability improvements proposed in the literature. 

\begin{figure*}
    \centering
    \includegraphics[width=\textwidth]{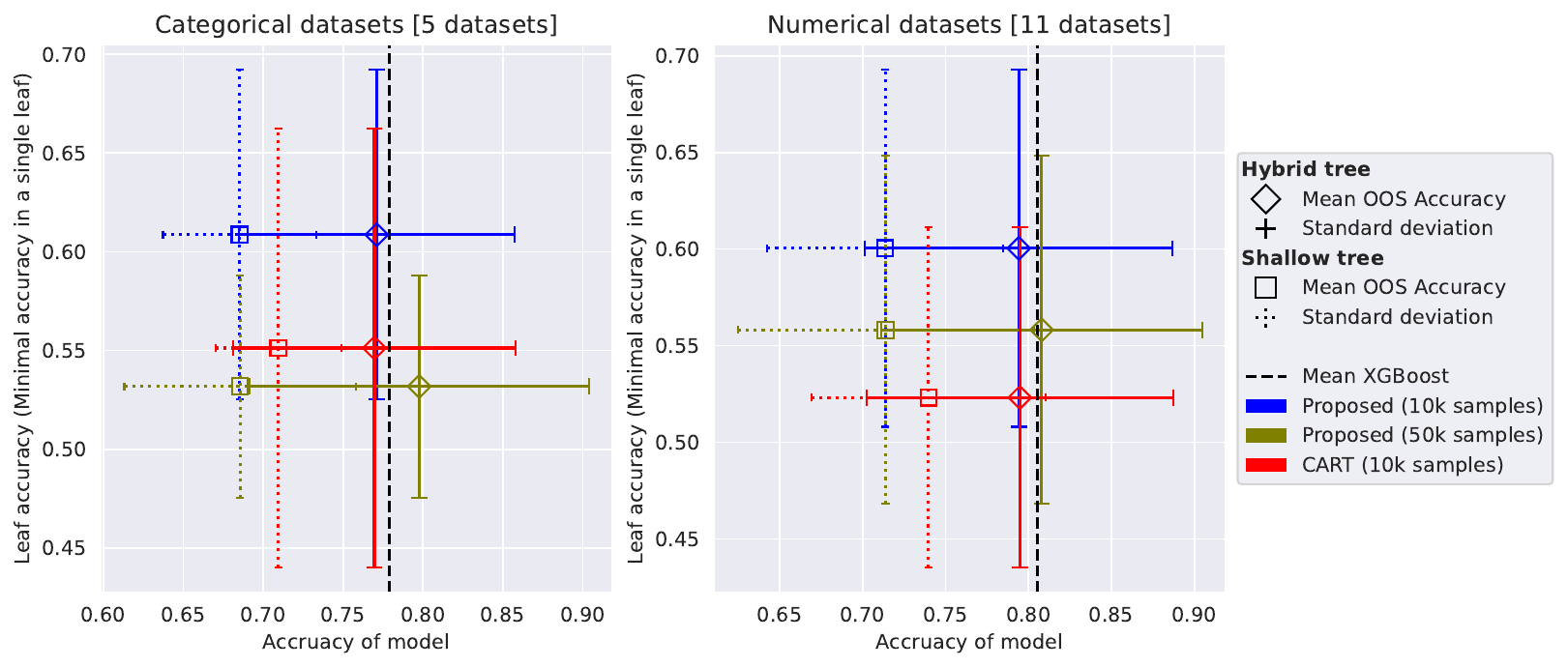}
    \caption{Comparison of a proposed model trained on at most 10,000 and 50,000 data samples. We use only datasets where the constraint caused a change, meaning we omit datasets with less than 12,500 samples. The number of datasets is in square brackets. All other presented models use only 10,000 samples, so the comparison of model accuracies is not fully representative.}
    \label{fig:compare50}
\end{figure*}

\subsection{More data}
The 10,000 size limit on training samples was suggested by the authors of the benchmark \citep{grinsztajnWhyTreebasedModels2022}. Another good reason for such a limit is that we want our model to balance the size of the formulation and the capability of the formulated model. In other words, if we take a small amount of data, we are less likely to grasp the intricacies of the target variable distribution within the dataset. And if we take too many samples, we create a formulation that will not achieve good performance in a reasonable time.

In a comparison of a model learned on a training dataset limited to 10,000 samples with a dataset limited to 50,000 samples, we see that more data does not necessarily lead to a better model, given the same time resources, see Figure \ref{fig:compare50}. The 50,000 model is worse because of the too-demanding complexity of the formulation. 

It improves the model accuracy, which is unsurprising since each leaf obtains more samples. The comparison to XGBoost is unreliable since the mean value for XGBoost was computed from the performance of models trained on at most 10,000 samples. 

\end{document}